\documentclass[journal]{IEEEtran}
\usepackage{graphicx}
\usepackage{latexsym}
\usepackage{float}
\usepackage{enumerate}
\usepackage{pdfpages}
\usepackage{amsmath}
\usepackage{amsfonts}
\usepackage{amssymb}
\usepackage[numbers,sort,compress]{natbib}
\usepackage{array}
\usepackage{booktabs}
\usepackage[font=small, labelfont=bf, skip=2pt]{caption}
\usepackage{tabularx}
\usepackage{subcaption}
\usepackage{footnote}
\usepackage{footmisc}

\usepackage{multirow}

\usepackage{siunitx}
\usepackage{xspace}
\usepackage{enumitem}

\makesavenoteenv{tabular}

\usepackage{color}

\newcommand{\hod}{{\tt hod}}
\newcommand{\hodg}{{\tt hodg}}
\newcommand{\jpd}{{\tt jpd}}
\newcommand{\jmv}{{\tt jmv}}

\makeatletter
\DeclareRobustCommand\onedot{\futurelet\@let@token\bmv@onedotaux}
\def\bmv@onedotaux{\ifx\@let@token.\else.\null\fi\xspace}
\def\eg{\emph{e.g}\onedot} 
\def\ie{\emph{i.e}\onedot} 
 
\def\etc{\emph{etc}\onedot}

\makeatother

\begin{document}
\bstctlcite{IEEEexample:BSTcontrol}
\title{A Comparative Review of Recent Kinect-based Action Recognition Algorithms}

\author{Lei~Wang,
        Du~Q.~Huynh,~\IEEEmembership{Senior Member,~IEEE}, Piotr Koniusz
\thanks{$\!\!\!\!\!\!\!$Lei Wang is pursuing a PhD degree in Computer Vision at the Australian National University (ANU) and Data61/CSIRO, Canberra, Australia, ACT 2601 (e-mail: lei.wang@data61.csiro.au). He is also associated with the University of Western Australia (UWA). 

Du Q. Huynh is with the Department of Computer Science and Software Engineering, the University of Western Australia,
Crawley, WA 6009, Australia (e-mail: du.huynh@uwa.edu.au).

Piotr Koniusz is with Data61/CSIRO (formerly known as NICTA) and the Australian National University, Canberra, Australia, ACT 2601 (e-mail: piotr.koniusz@data61.csiro.au).
} 
\thanks{$\!\!\!\!\!\!\!$Manuscript received on the 30\textsuperscript{th} of March, 2018, revised on the 19\textsuperscript{th} of April, 2019, accepted on the 24\textsuperscript{th} of June, 2019. Kindly cite the TIP rather than the ArXiV version.}}

\markboth{IEEE TRANSACTIONS ON IMAGE PROCESSING, ACCEPTED on the 24\textsuperscript{\MakeLowercase{th}} of June, 2019.}%
{Lei \MakeLowercase{\textit{et al.}}: A Comparative Review of Recent Kinect-based Action Recognition Algorithms}

\maketitle

\begin{abstract}
Video-based human action recognition is currently one of the most active research areas in computer vision. Various research studies indicate that the performance of action recognition is highly dependent on the type of features being extracted and how the actions are represented. Since the release of the Kinect camera, a large number of Kinect-based human action recognition techniques have been proposed in the literature. However, there still does not exist a thorough comparison of these Kinect-based techniques under the grouping of feature types, such as handcrafted versus deep learning features and depth-based versus skeleton-based features. In this paper, we analyze and compare ten recent Kinect-based algorithms for both cross-subject action recognition and cross-view action recognition using six benchmark datasets. In addition, we have implemented and improved some of these techniques and included their variants in the comparison. 
Our experiments show that the majority of methods perform better on cross-subject action recognition than cross-view action recognition, that skeleton-based features are more robust for cross-view recognition than depth-based features, and that deep learning features are suitable for large datasets.

\end{abstract}


\IEEEpeerreviewmaketitle


\section{Introduction}

\IEEEPARstart{H}{uman} action recognition has many useful applications such as human computer interaction, smart video surveillance, sports and health care. These applications are one of motivations behind much research work devoted to this area in the last few years. However, even with a large number of research papers found in the literature, many challenging problems, such as different viewpoints, visual appearances, human body sizes, lighting conditions, and speeds of action execution, still affect the performance of these algorithms. Further problems include partial occlusion of the human subjects by other objects in the scene and self-occlusion of human subjects themselves. Among the human action recognition papers presented in the literature, some of the early techniques focus on using conventional RGB videos~\cite{bobick2001, dollar2005, blank2005, ke2007, laptev2008, liu2008, bregonzio2009, liu2009}.
While these video-based techniques gave  promising results, their recognition accuracy is still relatively low, even when the scene is free of clutter. 



The Kinect camera introduced by Microsoft in 2001 was an attempt to broaden the 3D gaming experience of the Xbox 360's audience. However, as the Kinect camera can capture real-time RGB and depth videos, and there is a publicly available toolkit for computing the human skeleton model from each frame of a depth video, many research papers on 3D human action recognition  using the Kinect camera have emerged. One advantage of using depth videos than the conventional RGB videos is that it is easier to segment the foreground human subject even when the scene is cluttered. As depth videos do not have colour information, the colour of the clothes worn by the human subject has no effect on the segmentation process. This allows action recognition researchers to focus their effort more on getting robust feature descriptors to describe the actions rather than on low level segmentation. Numerous representative methods for 3D action analysis using depth videos include~\cite{Li2010, Yang2012, Oreifej2013, Rahmani2014, Rahmani2016, Rahmani2016CVPR}. These methods employ advanced machine learning techniques for which good results have been reported. Of course, depth images are also vulnerable to noise due to various factors~\cite{Mallick-et-al-Sensor14}. Thus,  using depth images does not always guarantee good action recognition performance~\cite{Li2010}.
The algorithm used for computing the 3D joint positions of the human skeletal model by the Kinect toolkit is based on the human skeleton tracking framework (OpenNI) of~\citet{Shotton2011}. In addition to the availability of the real-time depth video stream, this tracking framework also opens up the research area of skeleton-based human action recognition~\cite{Vemulapalli2014, amir2016, Vemulapalli2016, Kes2017, Ke2017, Rahmanids2017}. 


Human action recognition methods using the Kinect data can be classified into two categories, based on how the feature descriptors are extracted to represent the human actions. The first category is handcrafted features.  Action recognition methods using handcrafted features require two complex hand-design stages, namely \textit{feature extraction} and \textit{feature representation}, to build the final descriptor. Both the feature extraction stage and feature representation stage differ from one method to another. The feature extraction stage may involve computing the depth (and/or colour) gradients, histogram, and other more complex transformations of the video data. The feature representation stage may involve simple concatenation of the feature components extracted from the previous stage, a more complex fusion step of these feature components, or even using a machine learning technique, to get the final feature descriptor. 
These methods usually involve a number of operations that require researchers to carry out careful feature engineering and tuning.
Kinect-based human action recognition algorithms using handcrafted features reported in the literature include ~\cite{Li2010, Shotton2011, Yang2012, Oreifej2013, Rahmani2014, Rahmani2016, Vemulapalli2014, Vemulapalli2016, piotr2016, jingtian2018tip}. 

The second category is deep learning features. With the huge advance in neural network research in the last decade, deep neural networks have been used to extract high-level features from video sequences for many different applications, including 3D human action analysis. Deep learning methods reduce the need for feature engineering; however, they require a huge amount of labelled training data, which may not be available, and a long time to train. For small human action recognition datasets, deep learning methods may not give the best performance. Recent Kinect-based human action recognition algorithms are:~\cite{amir2016, Rahmani2016CVPR, Rahmanids2017, Rahmani2015, Kes2017, Ke2017, stgcn2018aaai, shuai2018, chaolong2018, hongsong2017, yansong2018, jun2017, zhiwu2017, inwoong2017, chenyang2018, hongsong2018, jun2018tpami, amor2018cvpr, nenggan2018aaai}. 

\smallskip

\noindent \textbf{Research contributions.}
Although both handcrafted and deep learning features have been used in human action recognition, 
to the best of our knowledge, a thorough comparison of recent action recognition methods for these two categories is not found in the literature. Our contributions in this paper are twofold: 
\begin{itemize}
\item We evaluate the performance of 10 recent state-of-art human action recognition algorithms, with specific focus on comparing the effectiveness of using handcrafted features versus deep learning features and skeleton-based features versus depth-based features. We believe that there is a lack of such a comparison in the literature on human action recognition. 
\item Furthermore, we evaluate the cross-view versus cross-subject performance of these algorithms and, for the multiview datasets, the impact of the camera view for both small and large datasets on human action recognition with respect to whether the features being used are depth-based, skeleton-based, or depth+skeleton-based. To the best of our knowledge, such evaluation has not been performed before. 
\end{itemize}

The paper is organized as follows. Section~\ref{sec:related_work} gives a brief review on recent human action recognition techniques. Section~\ref{sec:algorithms} covers the details of the 10 algorithms being compared in this paper. In Section~\ref{sec:experiments}, we describe our experimental setting and the benchmark datasets. Sections~\ref{sec:results} and \ref{sec:discussion} summarize our experimental results, comparison, and discussions. The last section concludes the paper.


\section{Related Work}
\label{sec:related_work}
\begin{table*}[t!]
\caption{Ten state-of-the-art action recognition methods evaluated in this paper.}
\begin{center}
\resizebox{\textwidth}{!}{\begin{tabular}{| l | l | l | l | l |}
\hline
 Algorithms & Year & Short descriptions & Kinect data used & Feature dimension\\ 
\hline
\hline
HON4D (Oreifej \& Liu)~\cite{Oreifej2013} & CVPR 2013 & handcrafted (global descriptor) & depth & $[17880, 151200]$\\
\hline
HDG (Rahmani et al.)~\cite{Rahmani2014} & WACV 2014 & handcrafted (local + global descriptor) & depth+skeleton & $[1662, 1819]$\\
\hline
LARP-SO (Vemulapalli \& Chellappa)~\cite{Vemulapalli2016} & CVPR 2016 & handcrafted (Lie Group) & skeleton & $3 \times 3 \times$ \#frames\\
\hline
HOPC (Rahmani et al.)~\cite{Rahmani2016} & TPAMI 2016 & handcrafted (local descriptor) & depth $\to$ pointcloud & depending on \#STKs$^\dagger$  \\
\hline
SCK+DCK (Koniusz et al.)~\cite{piotr2016} & ECCV 2016 & handcrafted (tensor representations) & skeleton & $\sim$ 40k \\
\hline
\hline
P-LSTM (Shahroudy et al.)~\cite{amir2016} & CVPR 2016 & deep learning (LSTM) & skeleton & \#joints ${\times 3 \times 8}^\ddagger$\\
\hline
HPM+TM (Rahmani \& Mian)~\cite{Rahmani2016CVPR} & CVPR 2016 & deep learning (CNN) & depth & 4096\\
\hline
Clips+CNN+MTLN (Ke et al.)~\cite{Ke2017} & CVPR 2017 & deep learning (pre-trained VGG19, MTLN) & skeleton & 7168 \\
\hline
IndRNN (Li et al.)~\cite{shuai2018} & CVPR 2018 & deep learning (independently RNN) & skeleton & 512
\\
\hline
ST-GCN (Yan et al.)~\cite{stgcn2018aaai} & AAAI 2018 & deep learning (Graph ConvNet) & skeleton & 256 \\
\hline

\end{tabular}}
\label{algorithms}
\end{center}
\vspace{-1ex}
$^\dagger$STK stands for \textit{spatio-temporal keypoint}.\quad
$^\ddagger$The P-LSTM features include 8 video segments, each of which is composed of a number of 3D joints.


\end{table*}

Action recognition methods can be classified into three categories based on the type of input data: colour-based~\cite{bobick2001, blank2005, dollar2005, fishkin2005, hodges2007,ke2007, liu2008, laptev2008, liu2009, bregonzio2009, buettner2009, Sung2011, Sung2012, Koppula2013,saurabh2014, hakan2016, amir2016pami, christoph2017, amlan2017, anoop2017, anoop2017cvpr, amir2017, joao2017, christoph2016cvpr, jianfang2018}, depth-based~\cite{Li2010, Yang2012,jiang2012, Oreifej2013, luxia2013, Rahmani2014, RahmaniHOPC2014, RahmaniLLC2014, yang2014, pichao2015,Rahmani2015, pichao2016, RahmaniPRL2016, Rahmani2016, Rahmani2016CVPR, amir2016pami, chenyang2017, Rahmanids2017, jianfang2018, zhiyuan2017, jingtian2018tip}, and skeleton-based~\cite{Shotton2011, Xia2012, Vemulapalli2014, Yong2015, vivek2015, wentao2016, jun2016, piotr2016, Vemulapalli2016, Amor2016, amir2016, Rahmanids2017, Ke2017, Kes2017, jianfang2018, hongsong2017, hongsong2018, chenyang2018, inwoong2017, zhiwu2017, jun2017, yansong2018, chaolong2018, jun2018tpami, amor2018cvpr, nenggan2018aaai, nour2018tip, jingtian2018tip, chenyang2019cvpr, maosen2019cvpr, lei2019cvpr1, lei2019cvpr2}. In this section, we will focus on reviewing recent methods using the last two types of features.

\vspace{0.05cm}
\noindent{\bf Depth-based action recognition.} 
Action recognition from depth videos ~\cite{Li2010, Yang2012,jiang2012, Oreifej2013,luxia2013, yang2014, RahmaniHOPC2014,RahmaniLLC2014, Rahmani2016,RahmaniPRL2016} has become more popular because of the availability of real-time cost-effective sensors. Most existing depth-based action recognition methods use global features such as space-time volume and silhouette information. 
For example, 
\citet{Oreifej2013}~captured the discriminative features by projecting the 4D surface normals obtained from the depth sequence onto a 4D regular space to build the \textit{Histogram of Oriented 4D Normals} (HON4D). \citet{yang2014} extended HON4D by concatenating local neighbouring hypersurface normals from the depth video to jointly characterize local shape and motion information. More precisely, they introduced an adaptive spatio-temporal pyramid to subdivide the depth video into a set of space-time cells for more discriminative features.
\citet{luxia2013} proposed to filter out the noise from the depth sensor so as to get more reliable spatio-temporal interest points for action recognition. Although these methods have achieved impressive performance for frontal 
action recognition, they are sensitive to changes of viewpoint. One way to alleviate this viewpoint issue is to directly process the pointclouds, as reported in the paper by \citet{Rahmani2016}. 

Apart from the methods mentioned above which use handcrafted features, the use of deep learning features~\cite{pichao2015, Rahmani2015, pichao2016,amir2016pami, Rahmani2016CVPR, Rahmanids2017, chenyang2017, zhiyuan2017} in human action recognition is on the rise. For example, \citet{pichao2015} used a \textit{Hierarchical Depth Motion Maps} (HDMMs) to extract the body shape and motion information and then trained a 3-channel deep \textit{Convolutional Neural Network} (CNN) on the HDMMs for human action recognition. 
In the following years, \citet{Rahmani2016CVPR} proposed to train a single \textit{Human Pose Model} (HPM) from real motion capture data to transfer the human pose from different unknown views to a view-invariant feature space, and \citet{chenyang2017} used a multi-stream deep neural networks to jointly learn the semantic relations among action attributes. 


\vspace{0.05cm}
\noindent{\bf Skeleton-based action recognition.} Existing skeleton-based action recognition methods can be grouped into two categories: joint-based methods and body part based methods. Joint-based methods model the positions and motion of the joints (either individual or a combination) using the coordinates of the joints extracted by the OpenNI tracking framework. For instance, a reference joint may be used and the coordinates of other joints are defined relative to the reference joint~\cite{Yang2012, Vemulapalli2014, Ke2017, Kes2017}, or the joint orientations may be computed relative to a fixed coordinate system and used to represent the human pose~\cite{Xia2012}, \etc For the body part based methods, the human body parts are used to model the human's articulated system. These body parts are usually modelled as rigid cylinders connected by joints. Information such as joint angles~\cite{Vemulapalli2016}, temporal evolution of body parts~\cite{Amor2016, amir2016,piotr2016, stgcn2018aaai}, and 3D relative geometric relationships between rigid body parts~\cite{Vemulapalli2014, Vemulapalli2016, stgcn2018aaai} has all been used to represent the human pose for action recognition. 

The method proposed by~\citet{Vemulapalli2014} falls into the body part based category. They represent the relative geometry between a pair of body parts, which may or may not be directly connected by a joint, as a point in $SE(3)$. Thus, a human skeleton is a point of the Lie group $SE(3) \times \cdots \times SE(3)$ where each action corresponds to a unique evolution of such a point in time. 
The approach of~\citet{Kes2017} relies on both body parts and body joints. The human skeleton model was divided into 5 body parts. A specific joint was selected for each body part as the reference joint and the coordinates of other joints were expressed as vectors relative to that reference joint. 
Various distance measures were computed from these vectors to yield a feature vector for each video frame.
The features vectors from all video frames were finally appended together and scaled to form a handcrafted greyscale image descriptor fed into a CNN. 
Somewhat related approach \cite{pk_bmvc18} uses kernels formed over body joints to obtain feature maps fed into a CNN for simultaneous action recognition and domain adaptation.

Recent human action recognition papers favour deep learning techniques to perform human action recognition. Apart from the CNN-based approaches~\cite{Kes2017,pk_bmvc18}, \textit{Recurrent Neural Networks} (RNNs) have also been popular~\cite{Yong2015, vivek2015, amir2016, wentao2016, jun2016, shuai2018, zhiyuan2017, hongsong2018, nenggan2018aaai}. Since \textit{Long Short-term Memory} (LSTM)~\cite{sepp1997} can model temporal dependencies as RNNs and even capture the co-occurrences of human joints, LSTM networks have also been a popular choice in human action recognition~\cite{jun2017, inwoong2017, amor2018cvpr, jun2018tpami, chenyang2019cvpr}. For instance, \citet{wentao2016} presented an end-to-end deep LSTM network with a \textit{dropout} step, \citet{amir2016} proposed a \textit{Part-aware Long Short-term Memory} (P-LSTM) network to learn the long-term patterns of the 3D trajectories for each grouped body part, and \citet{jun2018tpami} introduced the use of \textit{trust gates} in their spatio-temporal LSTM architecture.  

\vspace{0.05cm}
\noindent{\bf Action recognition via a combination of skeleton and depth features.} 
%
Combining skeleton and depth features together helps overcome situations when there are interactions between human subject and other objects or when the actions have very similar motion trajectories. Various action recognition algorithms~\cite{Rahmani2014, amir2016pami, Rahmanids2017, nour2018tip} that use both depth and skeleton features for robust human action recognition have been reported in recent years. For example, \citet{Rahmani2014}~proposed to combine 4 types of local features extracted from 
both depth images and 3D joint positions to deal with local occlusions and to increase the recognition accuracy. We refer to their method as \textit{HDG} from hereon. Another example is the approach of \citet{amir2016pami} where \textit{Local Occupancy Patterns} (LOP), HON4D, and skeleton-based features are combined with 
hierarchical mixed norms 
which regularize the weights in each modality group of each body part.
Recently, \citet{Rahmanids2017} used an end-to-end deep learning model to learn the body part representation from both skeletal and depth images. To improve the performance of the model, they adopted a bilinear compact pooling~\cite{yang2016cbp} layer for the generated depth and skeletal features. 
\citet{nour2018tip}, on the other hand, proposed to use canonical correlation analysis to maximize the correlation of features extracted from different sensors. The features investigated in their paper include bag of angles extracted from skeleton data, depth motion map from depth video, and optical flow from RGB video. The subspace shared by all the features was learned and average pooling was used to get the final feature descriptor.


\section{Analyzed and Evaluated Algorithms}
\label{sec:algorithms}
We chose ten action recognition algorithms shown in Table~\ref{algorithms} for our comparison and evaluation as they are recent action recognition methods (from 2013 onward) and they use skeleton-based, depth-based, handcrafted, and/or deep learning features. 
The technical details of these algorithms are summarized below.

\vspace{0.05cm}
\noindent{\bf HON4D}. \citet{Oreifej2013} presented a global feature descriptor that captures the geometry and motion of human action in the 4D space of spatial coordinates, depth and time. To form the HON4D descriptor, the space was quantized using a 600-cell polychoron with 120 vertices. The vectors stretching from the origin to these vertices were used as projection axes to obtain the distribution of normals for each video sequence. To improve the classification performance, random perturbations were added to those projectors.  
The dimensions of HON4D features (Table~\ref{algorithms}) vary across different datasets. 

\vspace{0.05cm}
\noindent{\bf HDG}. In this algorithm~\cite{Rahmani2014}, each depth sequence was firstly divided into small subvolumes; the histograms of depth and depth derivatives were computed for each subvolume. For each skeleton sequence, the torso joint was used as a stable reference joint for computing the histograms of joint position differences. In addition, the variations of each joint movement volume were incorporated into the global feature vector to form  spatio-temporal joint features. Two \textit{Random Decision Forests} (RDFs) were trained in this algorithm, one for feature pruning and one for classification. More details about feature dimensions of HDG and the feature pruning applied by us will be given in Section~\ref{sec:HDG-pruning}. 

\vspace{0.05cm}
\noindent{\bf HOPC}. Approach~\cite{Rahmani2016} models depth images as 3D pointclouds. The authors used two types of support volume, namely, so-called spatial support volume and spatio-temporal support volume. The HOPC descriptor was extracted from the pointcloud falling inside the support volume around each point, which may be classified as a \textit{spatio-temporal Keypoint} (STK) if the eigenvalue ratios of the pointcloud around it are larger than some predefined threshold. For each STK, the algorithm further projected  eigenvectors onto the axes of the 20 vertices of a regular dodecahedron. The final HOPC descriptor for each STK is a concatenation of 3 small histograms, each of which captures the distribution of an eigenvector of the pointcloud within the support volume.

\vspace{0.05cm}
\noindent{\bf LARP-SO}. \citet{Vemulapalli2016} extended their earlier work \cite{Vemulapalli2014} by the use of Lie Algebra Relative Pairs via $SO(3)$ for action recognition. We follow the convention adopted in~\cite{Rahmanids2017} and name this algorithm as LARP-SO. In this algorithm, the rolling map, which describes how a Riemannian manifold rolls over another one along a smooth rolling curve, was used for 3D action recognition. Each skeleton sequence was firstly represented by the relative 3D rotations between various human body parts, and each action was then modelled as a curve in the Lie Group. Since it is difficult to perform the classification of action curves in a non-Euclidean space, the curves were unwrapped 
by the logarithm map at a single point while a rolling map was used to reduce distortions. The \textit{Fourier Temporal Pyramid} (FTP) representation~\cite{jiang2012} was used in the algorithm to make the descriptor more robust to noise and less sensitive to temporal misalignments.


\vspace{0.05cm}
\noindent{\bf SCK+DCK}. \citet{piotr2016} used  tensor representations to capture the higher-order relationships between 3D human body joints for action recognition. They applied two different RBF kernels which they referred to as \textit{Sequence Compatibility Kernel} (SCK) and \textit{Dynamics Compatibility Kernel} (DCK). The former kernel  captures the spatio-temporal compatibility of joints while the latter  models the action dynamics of a sequence. An SVM was then trained on  linearized feature maps of such kernels for action classification.

\vspace{0.05cm}
\noindent{\bf HPM+TM}. Approach~\cite{Rahmani2016CVPR} employs a dictionary containing representative human poses from a motion capture database. A deep CNN architecture which is a modification of~\cite{saurabh2014} was then used to train a view-invariant human pose model. Real depth sequences were passed to the learned model frame-by-frame to extract high-level view-invariant features. 
Similarly to the LARP-SO algorithm above, the FTP was used to capture the temporal structure of the action videos.
The final descriptor for each video sequence is a collection of the Fourier coefficients from all the segments.

\vspace{0.05cm}
\noindent{\bf P-LSTM}. Approach~\cite{amir2016} proceeds by transforming 3D coordinates of the body joints from the camera coordinate system to the body coordinate system with the origin set at the spine. The 3D coordinates of all other body joints were then scaled based on the distance between the `hip centre' joint and the `spine' joint. A P-LSTM model was built by splitting the memory cells from the LSTM model into body part based sub-cells. For each video sequence, the pre-processed human joints were grouped into 5 parts (torso, two hands, and two legs) and the video was divided into 8 equal-sized video segments. Then, for a randomly selected frame per video segment, 3D coordinates of the joints inside each grouped part were concatenated and passed as input to the P-LSTM network to learn common temporal patterns of the parts and combine them into a global representation. 

\vspace{0.05cm}
\noindent{\bf Clips+CNN+MTLN}. \citet{Ke2017} presented a skeletal representation  referred to as \textit{clips}. 
The method proceeds by transforming the Cartesian coordinates of human joints (per skeleton sequence) into the cylindrical coordinates to generate 3 clips, with each clip corresponding to one channel of the cylindrical coordinates. To encode the temporal information for the whole video sequence, four stable joints (left shoulder, right shoulder, left hip and right hip) were selected as reference joints to produce 4 coordinate frames. The pre-trained VGG19 network~\cite{karen2015} was used as a feature extractor to learn the long-term spatio-temporal features from intermediate images formed from the 4 coordinate frames. Moreover, approach \cite{Ke2017} also employs the  \textit{Multi-task Learning Network} (MTLN) proposed by~\cite{ml1997} to incorporate the spatial structural information from the CNN features. 

\vspace{0.05cm}
\noindent{\bf IndRNN}. \citet{shuai2018} proposed a new RNN method, an Independently Recurrent Neural Network, for which neurons  per layer are independent of each other but they are reused across layers. Finally, multiple IndRNNs were stacked to build a deeper network than the traditional RNN.

\vspace{0.05cm}
\noindent{\bf ST-GCN}. The spatio-temporal graph representation for skeleton sequences proposed by \citet{stgcn2018aaai} is an extension of {\em Graph Convolutional Networks} (GCN) \cite{joan2014,defferrard2016, thomas2017} tailored to perform human action recognition. Firstly, the spatio-temporal graph is constructed
by inserting edges between neighbouring body joints (nodes) of the human body skeleton as well as along the temporal direction.
Subsequently, GCN and a classifier are applied to infer dependencies in the graphs (a single graph corresponds to a single action sequence) and perform classification.



\section{Experimental Setting}
\label{sec:experiments}

To perform experiments, we obtained off-the-shelf codes for HON4D~\cite{Oreifej2013}, HOPC~\cite{Rahmani2016}, LARP-SO~\cite{Vemulapalli2016}, HPM+TM~\cite{Rahmani2016CVPR}, IndRNN~\cite{shuai2018} and ST-GCN~\cite{stgcn2018aaai} from the respective authors' websites. 
For SCK+DCK~\cite{piotr2016}, HDG~\cite{Rahmani2014},  P-LSTM~\cite{amir2016} and Clips+CNN+MTLN~\cite{Ke2017},
we used our own Matlab implementations given that codes for these methods are not publicly available. 
Moreover, we employed ten variants of the HDG~\cite{Rahmani2014} representation so as to evaluate the performance with respect to different combinations of its individual descriptor types. We also implemented the traditional RNN and LSTM as baseline methods, and added four variants of P-LSTM to evaluate the impact of using different numbers of video segments for skeletal representation as well as different numbers of hidden neurons. 

\subsection{Benchmark Datasets}

Listed in Table \ref{datasets} are six benchmark datasets used in our evaluation, each of which is detailed below.

\begin{table*}[t!]
\caption{Six publicly available benchmark datasets used in our experiments for 3D action recognition.}
\begin{center}
\resizebox{\textwidth}{!}{\begin{tabular}{| l | c | c | c | c | c | l | c | c |}
\hline
 Datasets & Year & Classes & Subjects & \#Views & \#videos & Sensor & Modalities & \#joints \\ 
\hline
\hline
MSRAction3D~\cite{Li2010} & 2010 & 20 & 10 & 1 & 567 & Kinect v1 & Depth + 3DJoints & 20\\
\hline
3D Action Pairs~\cite{Oreifej2013} & 2013 & 12 & 10 & 1 & 360 & Kinect v1 & RGB + Depth + 3DJoints & 20\\
\hline
CAD-60~\cite{Sung2011} & 2011 & 14 & 4 & -- & 68 & Kinect v1 & RGB + Depth + 3DJoints & 15\\
\hline
UWA3D Activity Dataset~\cite{RahmaniHOPC2014} & 2014 & 30 & 10 & 1 & 701 & Kinect v1 & RGB + Depth + 3DJoints & 15\\
\hline
UWA3D Multiview Activity II~\cite{Rahmani2016} & 2015 & 30 & 9 & 4 & 1070 & Kinect v1 & RGB + Depth + 3DJoints & 15\\
\hline
NTU RGB+D Dataset~\cite{amir2016} & 2016 & 60 & 40 & 80 & 56880 & Kinect v2 & RGB + Depth + IR + 3DJoints & 25\\
\hline
\end{tabular}}
\label{datasets}
\end{center}
\vspace{-1ex}
{\footnotesize{(The number of views is not stated in the CAD-60 dataset.)}}
\end{table*}

\vspace{0.05cm}
\noindent{{\bf MSRAction3D}}~\cite{Li2010} is one of the earliest action datasets captured with the Kinect depth camera. It contains 20 human sport-related activities such as {\it jogging}, {\it golf swing} and {\it side boxing}. Each action in this dataset was performed 2 or 3 times by 10 people. This dataset is challenging because of high inter-action similarities.

\vspace{0.05cm}
\noindent{{\bf 3D Action Pairs}}~\cite{Oreifej2013} contains 6 selected pairs of actions that have very similar motion trajectories, \eg, {\it put on a hat} and {\it take off a hat}; {\it pick up a box} and {\it put down a box}; {\it stick a poster} and {\it remove a poster}. Each action was performed 3 times by 10 people. There are two challenging aspects of this dataset: (i)~the actions in each pair have similar motion trajectories; (ii)~the object that is interacted by the subject in each video is only present in the RGB-D data but not the skeleton data.

\begin{figure}[t!]
    \centering 
\begin{subfigure}{0.3\columnwidth}
  \includegraphics[width=\linewidth]{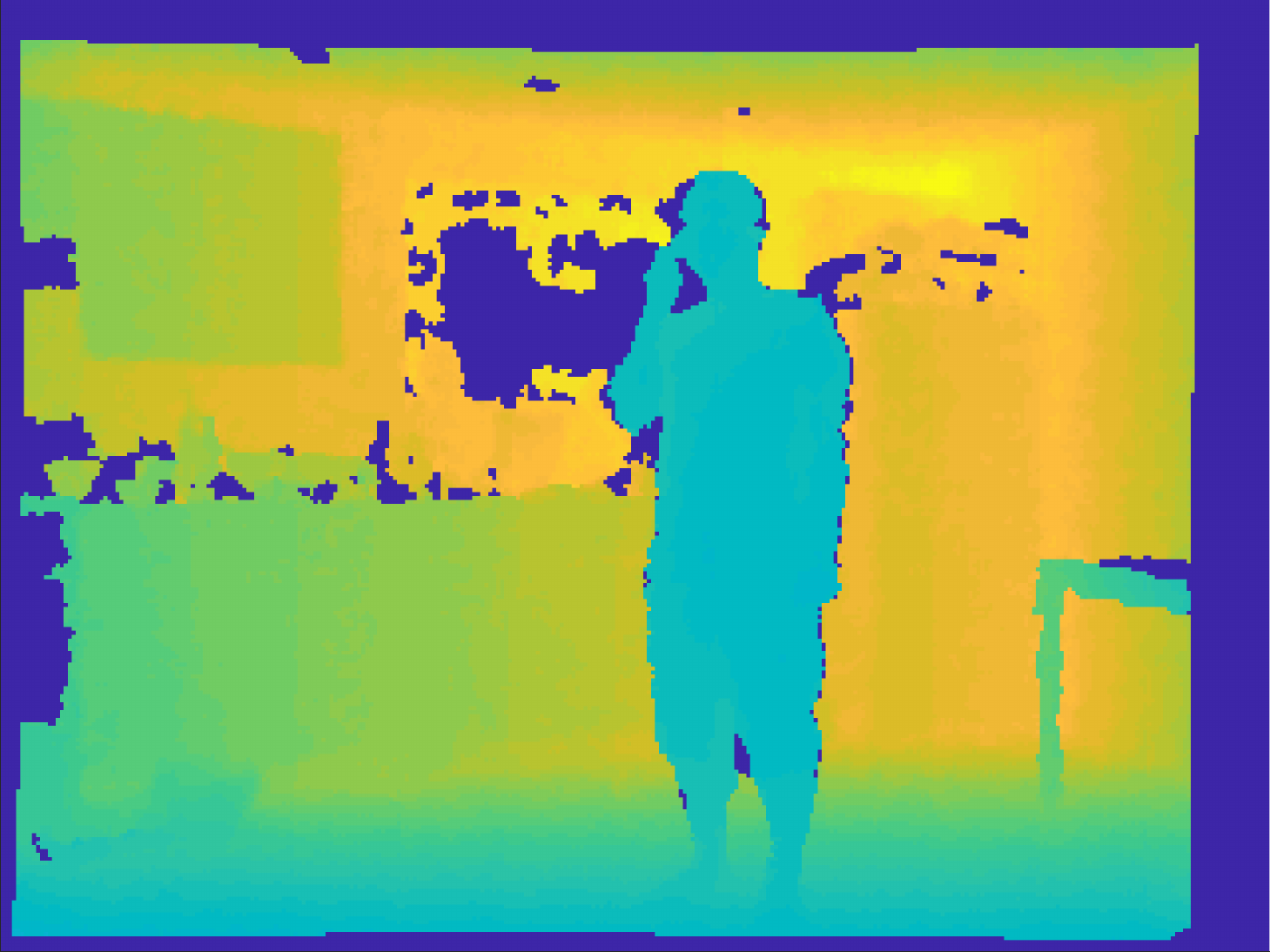}
  \caption{talking (phone)}
  \label{fig:1}
\end{subfigure}\hfil 
\begin{subfigure}{0.3\columnwidth}
  \includegraphics[width=\linewidth]{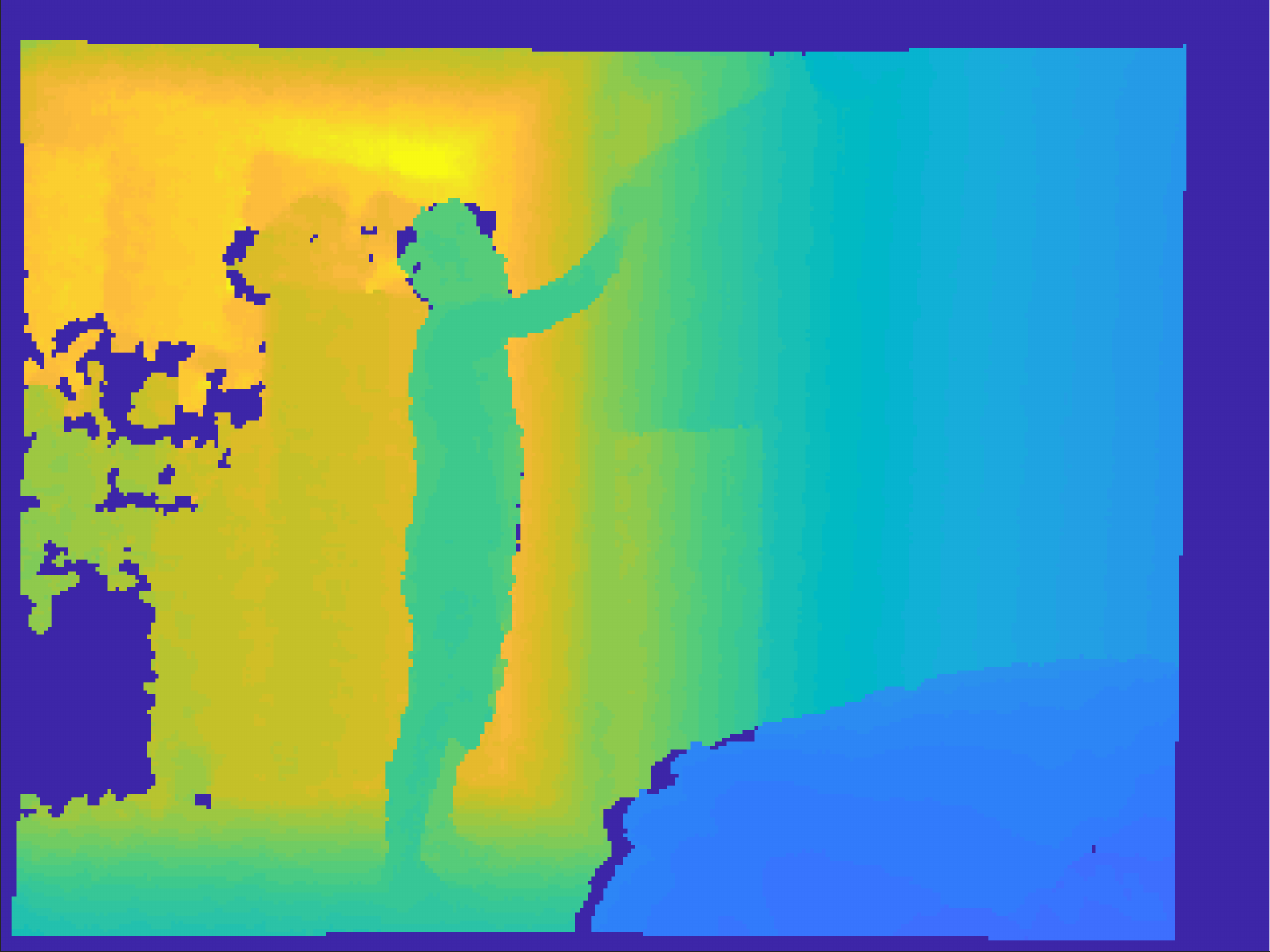}
  \caption{writing}
  \label{fig:2}
\end{subfigure}\hfil 
\begin{subfigure}{0.3\columnwidth}
  \includegraphics[width=\linewidth]{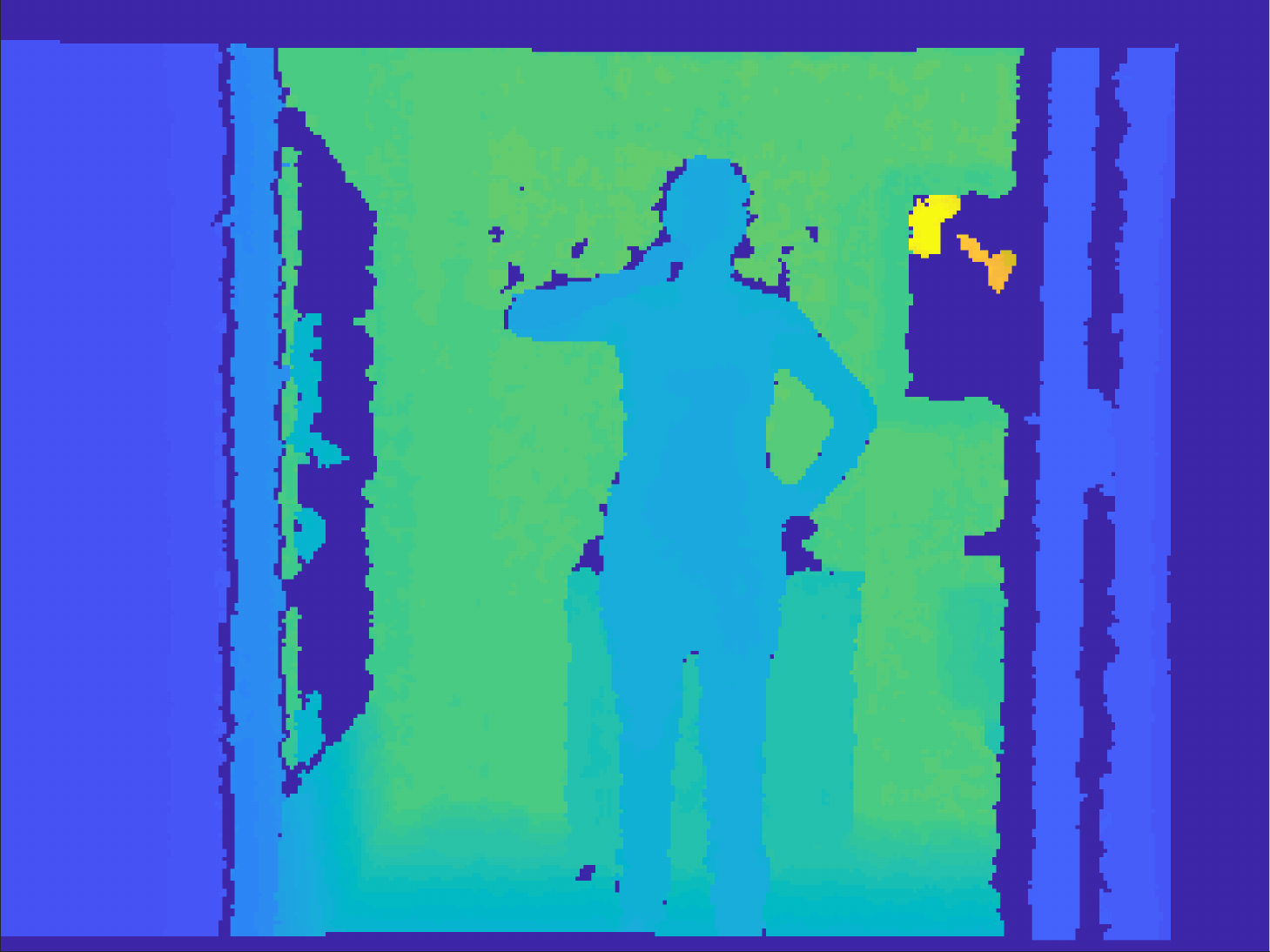}
  \caption{brushing teeth}
  \label{fig:3}
\end{subfigure}
\medskip
\begin{subfigure}{0.3\columnwidth}
  \includegraphics[width=\linewidth]{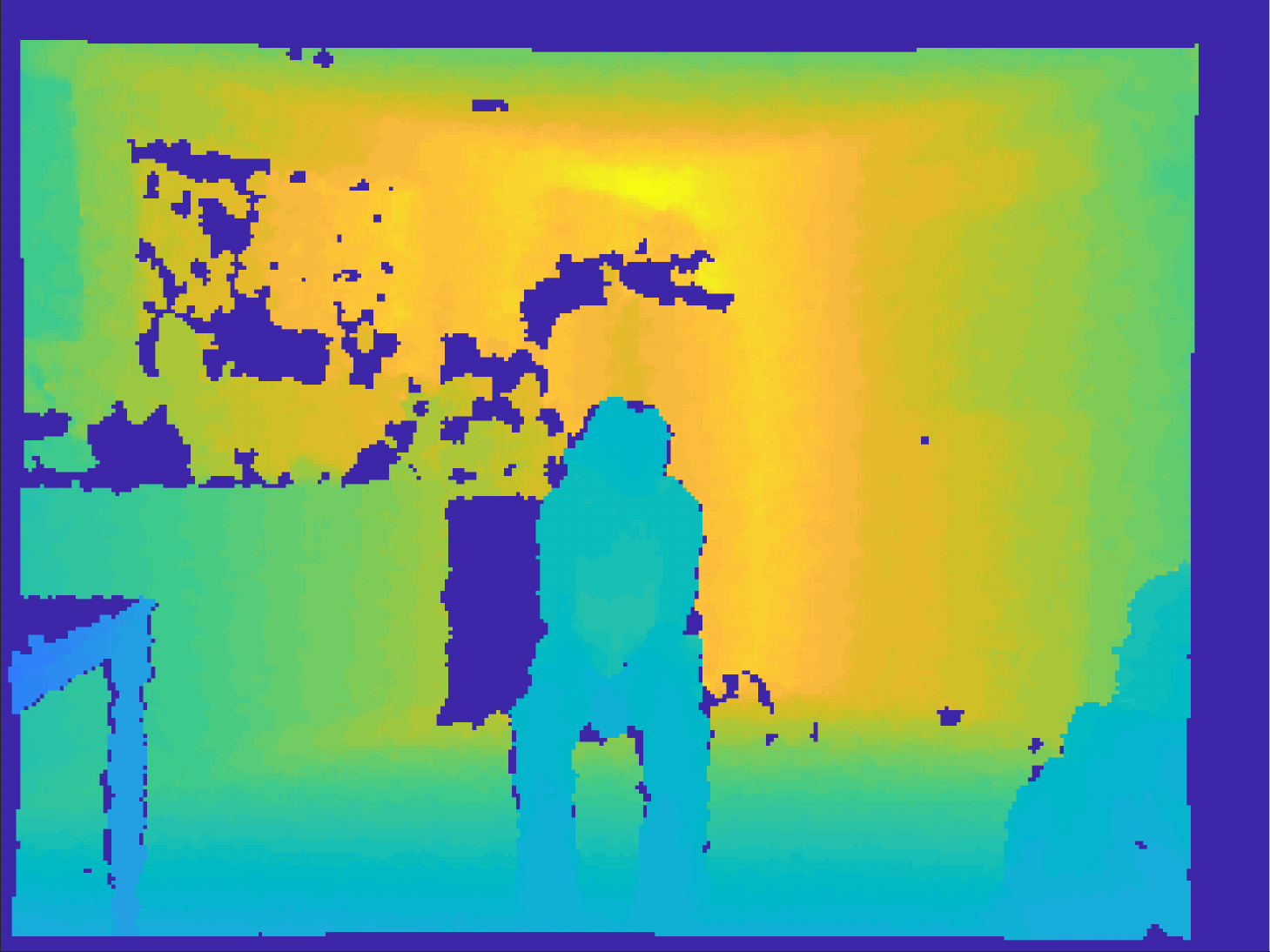}
  \caption{talking (couch)}
  \label{fig:4}
\end{subfigure}\hfil
\begin{subfigure}{0.3\columnwidth}
  \includegraphics[width=\linewidth]{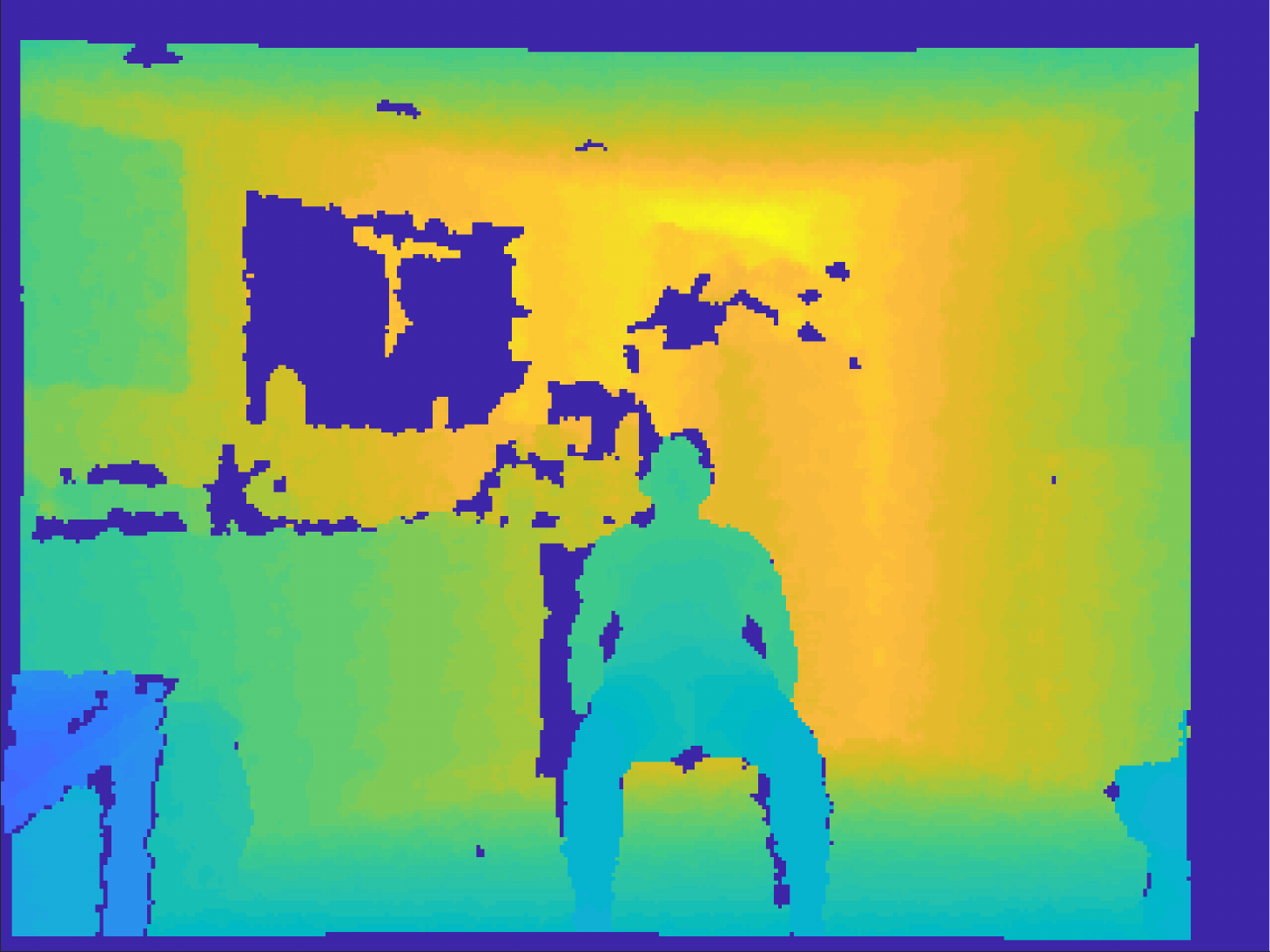}
  \caption{relaxing (couch)}
  \label{fig:5}
\end{subfigure}\hfil 
\begin{subfigure}{0.3\columnwidth}
  \includegraphics[width=\linewidth]{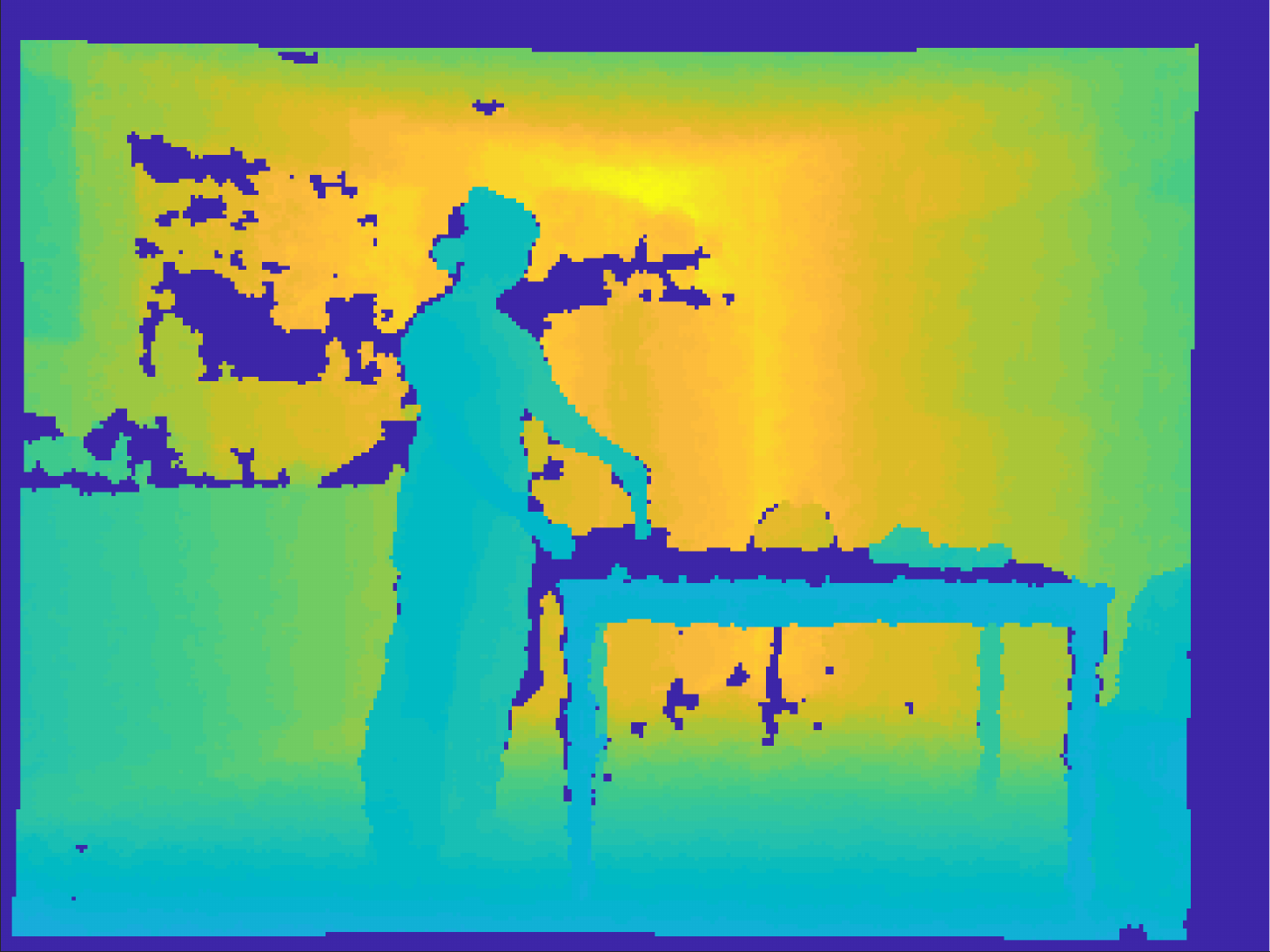}
  \caption{cooking (stirring)}
  \label{fig:6}
\end{subfigure}
\caption{Sample depth images from the CAD-60 dataset.}
\label{cad60examples}
\end{figure}

\vspace{0.05cm}
\noindent{{\bf Cornell Activity Dataset} (CAD)}~\cite{Sung2011} comprises two sub-datasets, CAD-60 and CAD-120. Both sub-datasets contain RGB-D and tracked skeleton video sequences of human activities captured by a Kinect sensor. In this paper, only CAD-60 was used in the experiments. Fig.~\ref{cad60examples} illustrates depth images from the CAD-60 dataset and demonstrates that this dataset exhibits high levels of noise in its depth videos.

\vspace{0.05cm}
\noindent{{\bf UWA3D Activity Dataset}}~\cite{RahmaniHOPC2014} contains 30 actions performed by 10 people of various height at different speeds in cluttered scenes. 
This dataset has high between-class similarity and contains frequent self-occlusions.

\begin{figure}[t!]
\centering
\includegraphics[width=\linewidth]{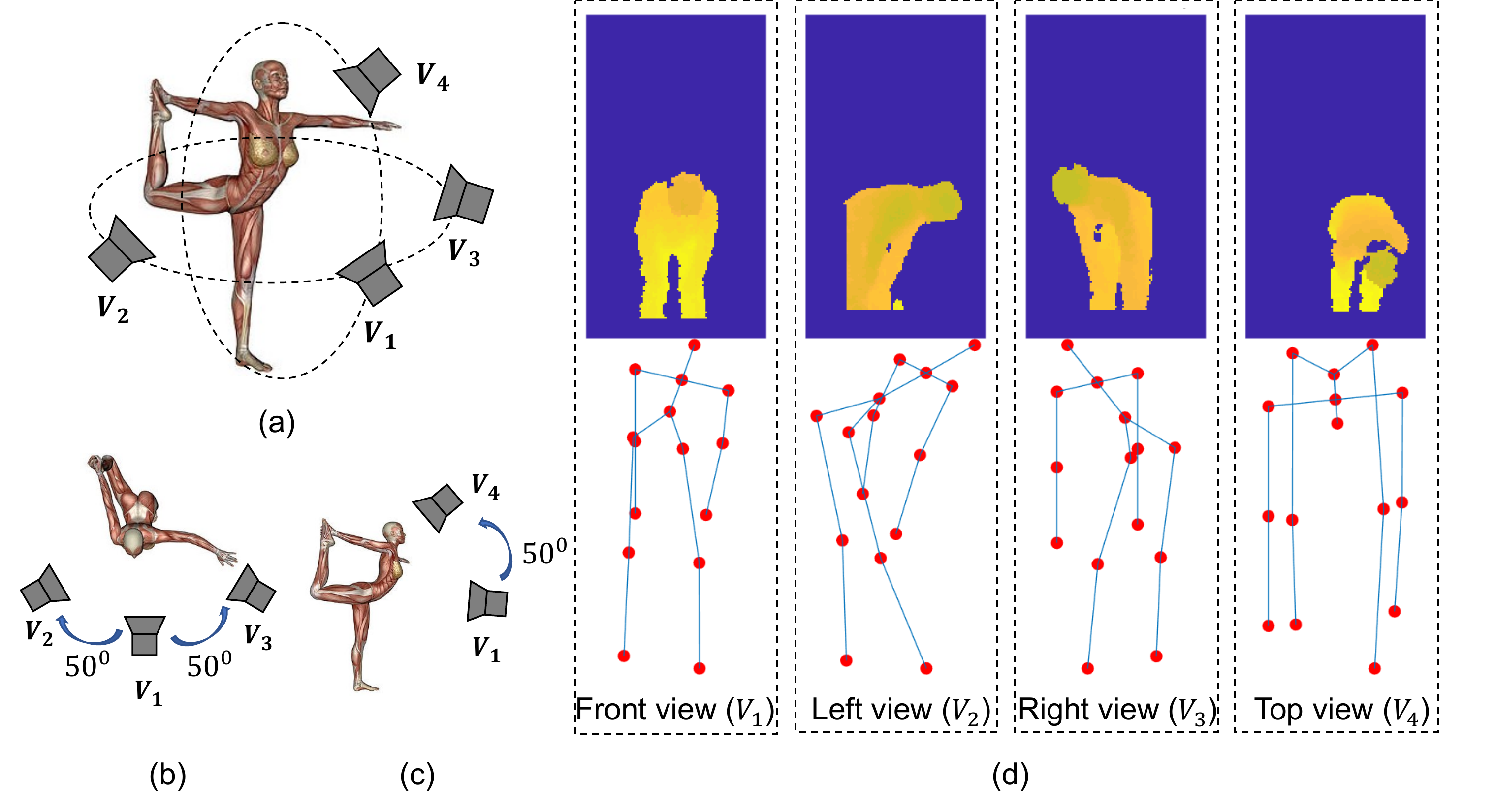}
\caption{(a) A perspective view of the camera setup in the UWA3D Multiview Activity II dataset. The views $V_1$, $V_2$ and $V_3$ are at the same height. (b) and (c) show the top and side views of the setup. The angles between $V_1$ and $V_2$, between $V_1$ and $V_3$, and between $V_1$ and $V_4$ are all approximately 50 degrees~\cite{RahmaniHOPC2014}. (d) An example video frame of the depth and skeleton data for the {\it bending} action.}
\label{uwa3d_sample}
\end{figure}

\vspace{0.05cm}
\noindent{{\bf UWA3D Multiview Activity II}}~\cite{Rahmani2016} contains 30 actions performed by 9 people in a cluttered environment. In this dataset, the Kinect camera was moved to different positions to capture the actions from 4 different views (see Fig.~\ref{uwa3d_sample}(a)-(c)): front view ($V_1$), left view ($V_2$), right view ($V_3$), and top view ($V_4$). This dataset is therefore more challenging than the previous four datasets. Fig.~\ref{uwa3d_sample}(d) shows sample video frames from this dataset.

\vspace{0.05cm}
\noindent{{\bf NTU RGB+D Dataset}}~\cite{amir2016} is so far the largest Kinect-based action dataset which contains 56,880 video sequences and over 4 million frames. There are 60 action classes performed by 40 subjects captured from 80 views with 3 Kinect v.2 cameras. 
This dataset has variable sequence lengths for different sequences and exhibits high intra-class variations.

\vspace{0.05cm}
\noindent{{\bf Dataset usage.}} Below we detail how the above six datasets were used in our experiments.

The MSRAction3D, 3D Action Pairs, CAD-60 and UWA3D Activity datasets were used for cross-subject (single-view) experiments. For every dataset, we used half of the subjects' data for training and the remaining half for testing. We tested all the possible combinations of subjects for the training and testing splits to obtain the average recognition accuracy of each algorithm. For example, for 10 subjects in the MSRAction3D dataset, $\binom{10}{5}=252$ experiments were carried out.

The UWA3D Multiview Activity II dataset was used for cross-view experiments, with two views of the samples being used for training and the remaining views for testing. There were 12 different view combinations in the experiments.

The NTU RGB+D dataset was used in both cross-subject and cross-view experiments. Despite indications that this dataset has 80 views of human action recognition, the data samples were grouped according to three camera sets. For cross-view action recognition, we used the video sequences captured by two cameras as our training data and the remaining sequences for testing. A total of 3 different camera combinations were experimented with.
 
\subsection{Evaluation Settings}


Below we detail the experimental settings of the algorithms.

\vspace{0.05cm}
\noindent{{\bf HON4D}}. According to~\cite{Oreifej2013}, HON4D  has the frame size of $320 \times 240$ and each video is divided into $4 \!\times\! 3 \!\times\! 3$ (${\text {width}} \!\times\! {\text {height}} \!\times\! {\text {\#frames}}$) spatio-temporal cells. In our evaluations, we used these same settings for all the datasets.

\vspace{0.05cm}
\noindent{{\bf HOPC}}. \citet{Rahmani2016} used different spatial and temporal scales for different datasets. In this paper, a constant temporal scale and spatial scale were used for all the datasets. For the MSRAction3D and 3D Action Pairs datasets, the temporal and spatial scales were set to 2 and 19, respectively. For the remaining datasets, we used 2 for the temporal scale and 140 as the spatial scale. Moreover, we divided each depth video into $6 \!\times\! 5 \!\times\! 3$ spatio-temporal cells (along the $X$, $Y$ and {\it time} axes) to extract features.

\vspace{0.05cm}
\noindent{{\bf LARP-SO}}. The desired number of frames~\cite{Vemulapalli2016} used for computing skeletal representation varies depending on the datasets used in the experiments. The desired frame numbers for the UWA3D Activity, UWA3D Multiview Activity II, and NTU RGB+D datasets were all set to 100. For the MSRAction3D, 3D Action Pairs datasets, and CAD-60, they were set to 76, 111, and 1,000, respectively.

\vspace{0.05cm}
\noindent{{\bf SCK+DCK}}. We followed the experimental settings described in~\cite{piotr2016,piotr2019} for all the datasets and we used authors' newest model which aggregates over subsequences (not just sequences). For SCK, we normalized all human body joints with respect to the hip joints across frames as well as the lengths of all body parts. For DCK, we used the unnormalized body joints, and assumed that the displacements of body joint coordinates across frames captured their temporal evolution.

\vspace{0.05cm}
\noindent{{\bf HDG}}. According to~\cite{Rahmani2014}), the number of used subvolumes  has no significant effect to the discriminative features, we divided each video sequence into $10 \!\times\! 10 \!\times\! 5$ subvolumes (along $X$, $Y$ and {\it time}) for computing the histograms of depth as well as the depth gradients. For the joint movement volume features, each joint volume was divided into  $1 \!\times\! 1 \!\times\! 5$ cells (along $X$, $Y$ and {\it time}). There are four individual feature representations encapsulated by HDG: 
\renewcommand{\labelenumi}{(\roman{enumi})}
\begin{enumerate}[leftmargin=0.8cm]
\item histogram of depth ({\hod}), 
\item histogram of depth gradients ({\hodg}),
\item joint position differences ({\jpd}), 
\item joint movement volume features ({\jmv}).
\end{enumerate}
We follow \cite{lei2017} and evaluate the performance of the 10 variants of HDG in our experiments.

\vspace{0.05cm}
\noindent{{\bf HPM+TM}}. We followed~\cite{Rahmani2016CVPR} and set the number of Fourier Pyramid levels to 3 and the number of low frequency Fourier coefficients to 4 for all datasets. We used the human pose model trained by~\citet{Rahmani2016CVPR} to extract view-invariant features from each depth sequence. We also compared the recognition accuracies of this algorithm given Average Pooling (AP) versus Temporal Modelling (TM) used for extraction of CNN features.

\vspace{0.05cm}
\noindent{{\bf P-LSTM}}. We applied the same normalization preprocessing step as in~\cite{amir2016} for the skeletal representation. In our experiments, the number of video segments and the number of hidden neurons for 1-layer RNN, 2-layer RNN, 1-layer LSTM, 2-layer LSTM, 1-layer P-LSTM and 2-layer P-LSTM were all set to 8 and 50, respectively. We also evaluated the performance of different numbers of video segments and different numbers of hidden neurons in P-LSTM. The learning rate and the number of epochs in our experiments were set to 0.01 and 300, respectively.

\vspace{0.05cm}
\noindent{{\bf Clips+CNN+MTLN}}. The learning rate was set to 0.001 and the batch size was set to 100 for MTLN. We selected four different experimental settings from \cite{Ke2017} to compare the performance of recognition: Frames+CNN, Clips+CNN+Pooling, Clips+CNN+Concatenation, and Clips+CNN+MTLN.

\vspace{0.05cm}
\noindent{{\bf IndRNN}}. We used the Adam optimizer with the initial learning rate $2 \times 10^{-4}$ and applied the decay by 10 once the evaluation accuracy did not increase. For cross-subject and cross-view experiments, the dropout rates were set to 0.25 and 0.1, respectively.


\vspace{0.05cm}
\noindent{{\bf ST-GCN}}. For the convolution operations, we used the optimal partition strategy according to the ablation study in~\cite{stgcn2018aaai}. As different datasets have different numbers of body joints (see Table~\ref{datasets}), we reconstructed the spatio-temporal skeleton graphs. For NTU RGB+D dataset, we used the same experimental settings as described in~\cite{stgcn2018aaai} (\eg, we work with up to two human subjects per sequence). For the remaining 5 datasets, we used a different setting as only one performing subject was present per video.

Moreover, we performed extra experiments for IndRNN and ST-GCN: instead of using 3D skeleton sequences as inputs, we used {\jpd} features which redefine the  3D skeleton joint positions by translating them to be centred at the torso (or `spine') joint.

\subsection{Evaluation Measure}

The recognition accuracy $A_c$ of an algorithm for any given action class $c$ is defined as the proportion of correct class $c$ labels returned by the algorithm:
\begin{equation}
A_c=\text{\#correct\_class\_$c$\_labels}\, \left/ \,\text{\#actual\_class\_$c$\_labels}\right..
\label{eq:accuracy}
\end{equation}

\newcommand{\Ntrees}{{N_{\text{trees}}}}

To show the recognition accuracies of an algorithm for all the action classes, a confusion matrix is often used.
The overall performance of an algorithm on a given dataset is evaluated using the average recognition accuracy $\bar{A}$ defined
as:
$\bar{A}=\frac{1}{C}\sum_{c=1}^{C}{A_c}$,
where $C$ is the total number of action classes in a given dataset.

To show the overall performance of each algorithm on $M$ datasets, we first rank the performance of each algorithm from 1 to 5 (a lower rank value represents a better performance) based on the recognition accuracy so each algorithm has its own rank value $r_i$ given the $i$\textsuperscript{th} dataset. We then compute the Average Rank (AVRank) as follows:

\begin{equation}
\text{AVRank}=\frac{1}{M}{\textstyle\sum}_{i=1}^{M}{r_i}.
\label{eq:rank}
\end{equation}

\subsection{Optimisation of Hyperparameters for HDG}
\label{sec:HDG-pruning}
There are 3 hyperparameters in the HDG algorithm. The first hyperparameter is the number of subvolumes, which we set to the same value as in~\cite{Rahmani2014}. The second and third hyperparameters, which are the number of trees $\Ntrees$ used in training and the threshold $\theta$ used in feature pruning, were optimised during our experiments (Table~\ref{tab:HDG-pruned}). As the length of the combined features in the HDG algorithm is large (\eg, the length of the HDG-all features for the MSRAction3D dataset is 13,250), we trained one RDF to select the feature components that have high importance values. This helped to increase the processing speed without compromising  the recognition accuracy.

We evaluated the effect of hyperparameters $\Ntrees$ and $\theta$ on the HDG algorithm for different combinations of individual HDG features using the MSRAction3D dataset (for single-view) and the UWA3D Multiview Activity II dataset (for cross-view). 
Table~\ref{tab:HDG-pruned} shows the optimal values for $\Ntrees$ and $\theta$ obtained from the grid search. The corresponding dimensions of different HDG combined features before and after pruning are also indicated. Compared to other individual features of HDG, {\jpd} and {\jmv} are small-sized features, thus when used alone in both datasets, their dimensions are not reduced by much during feature pruning. However, when either or both of them are combined with other individual features in cross-view action recognition, their importance values are significantly higher. This allows a large reduction in feature dimension after pruning (see the last 4 rows of the table). Our experiments in the next section also confirm that skeleton-based features deal better with the view-invariance than depth-based features.


The optimal values of $\Ntrees$ and $\theta$ shown in Table~\ref{tab:HDG-pruned} were used to prune the HDG features for all datasets. As different datasets have different numbers of body joints (see Table~\ref{datasets}), the dimensions of these HDG features after pruning across datasets are not the same.


\begin{table*}[tbp!]
\caption{Optimal hyperparameter values and feature dimensions before and after pruning for the HDG combined features.}
\label{tab:HDG-pruned}
\begin{center}
\resizebox{\textwidth}{!}{\begin{tabular}{
|l||r|S[table-format=2.2]|c|r||r|
S[table-format=2.2]|c|r|}
\hline
Combination & \multicolumn{4}{c||}{MSRAction3D} & \multicolumn{4}{c|}{UWA3D Multiview Activity II} \\
\cline{2-9}
of individual & dimensions & {\text{~~optimal $\theta$~~}} & optimal $\Ntrees$ & dimensions & dimensions & {\text{~~optimal $\theta$~~}} & optimal $\Ntrees$ & dimensions \\
features & before pruning & {\text{$\times 10^{-3}$}} & & after pruning & 
before pruning & {\text{$\times 10^{-3}$}} & & after pruning \\
\hline\hline
HDG-{\hod} & 2,500 & 1.5 & 100 & 1,442  & 2,500 & 2.7 & 60 &  1,231  \\
HDG-{\hodg} & 10,000 & 20.9 & 200 &  550  & 10,000 & 2.4 & 60 & 3,891  \\
HDG-{\jpd} & 150 & 11.5 & 80 & 148  & 150 & 48.1 & 120 &  142 \\
HDG-{\jmv} & 600 & 3.4 & 140 & 571  & 450 & 3.1 & 60 &   449 \\
HDG-{\hod}+{\hodg} & 12,500 & 17.0 & 180 &  786  & 12,500 & 4.7 & 140 &  3,987 \\
HDG-{\jpd}+{\jmv} & 750 & 2.0 & 80 & 690  & 600 & 6.8 & 100 &  581 \\
HDG-{\hod}+{\hodg}+{\jpd}  & 12,650 & 8.2 & 160 &  2,221  & 12,650 & 29.4 & 80 & 239  \\
HDG-{\hod}+{\hodg}+{\jmv}  & 13,100 & 7.4 & 180 &  2,189  & 12,950 & 25.0 & 100 & 135  \\
HDG-{\hodg}+{\jpd}+{\jmv} & 10,750 & 8.3 & 180 & 1,711  & 10,600 & 15.2 & 140 &  456 \\
HDG-all features & 13,250 & 13.3 & 120 & 1,013  & 13,100 &  19.0 & 100 & 300 \\
\hline
\end{tabular}}
\end{center}



\end{table*}

\section{Experimental Results}
\label{sec:results}
\subsection{MSRAction3D, 3D Action Pairs, CAD-60, and UWA3D Activity Datasets}

\begin{table*}[t!]
\caption{Comparison of average cross-subject action recognition accuracies (percentage) for the four single-view datasets (\ie, $M=4$ in Eq.~\eqref{eq:rank}). Each block of rows shows the performance of one method and its variants. 
The best algorithm for each dataset is highlighted in bold. The last column of the table shows the average rank of the best performing algorithm in each block. The final rank values are computed using Eq.~\eqref{eq:rank}, where top performing methods have smaller rank values. Other poorer performing methods in the same block are not considered for their rank values, so their final ranks are marked as `--'.}
\begin{center}
\resizebox{\textwidth}{!}{\begin{tabular}{ c | l | c | c | c | c | c |}
\cline{1-7}
\multicolumn{1}{c|}{~} & {\bf Method} & {\bf MSRAction3D} & {\bf 3D Action Pairs} & {\bf CAD-60} &
{\bf UWA3D Activity} & {\bf AVRank}\\ 
\hline
\multirow{16}{*}{\parbox{1.0cm}{Hand-crafted features}}
& HON4D~\cite{Oreifej2013} (Depth) & 82.15 & {\bf 96.00} & 72.70 & 48.89 & 3.25\\
\cline{2-7}
& HOPC~\cite{Rahmani2016} (Depth)  & 85.49 & 92.44 & 47.55 &  60.58 & 3.25\\
\cline{2-7}
& LARP-SO-logarithm map \cite{Vemulapalli2016} (Skel.) & 88.69 & 92.96 & 69.12 & 51.96 & -- \\ 
& LARP-SO-unwrapping while rolling \cite{Vemulapalli2016} (Skel.) & 88.47 & 94.09 & 69.12 & 53.05 & -- \\
& LARP-SO-FTP \cite{Vemulapalli2016} (Skel.) & 89.40 & 94.67 & 76.96 & 50.41 & 2.50\\ 
\cline{2-7}
& HDG-{\hod}~\cite{lei2017} (Depth) & 66.22 & 81.20 & 26.47 & 44.35 & --\\
& HDG-{\hodg}~\cite{lei2017} (Depth) & 70.34 & 90.98 & 50.98 & 54.23 & --\\ 
& HDG-{\jpd}~\cite{lei2017} (Skel.) & 55.54 & 53.78 & 46.08 & 40.88 & -- \\
& HDG-{\jmv}~\cite{lei2017} (Skel.) & 62.40 & 84.87 & 41.18 & 51.02 & --\\ 
& HDG-{\hod}+{\hodg}~\cite{lei2017} (Depth) & 71.81 & 90.96 & 51.96 & 55.17 & --\\
& HDG-{\jpd}+{\jmv}~\cite{lei2017} (Skel.) & 65.57 & 84.93 & 49.02 & 55.57 & --\\
& HDG-{\hod}+{\hodg}+{\jpd}~\cite{lei2017} (Depth + Skel.) & 72.06 & 90.72 & 51.47 & 56.41 & --\\
& HDG-{\hod}+{\hodg}+{\jmv}~\cite{lei2017} (Depth + Skel.) & 75.41 & 92.27 & 49.51 & 58.80 & --\\
& HDG-{\hodg}+{\jpd}+{\jmv}~\cite{lei2017} (Depth + Skel.) & 75.00 & 92.28 & 52.94 & 59.82 & --\\
& HDG-all features~\cite{lei2017} (Depth + Skel.) & 75.45 & 92.13 & 51.96 & 60.33 & 4.00\\
\cline{2-7}
& SCK+DCK~\cite{piotr2016} (Skel.) & {\bf 89.47} & {\bf 96.00} & {\bf 89.22} & {\bf 61.52} & 1.00 \\
\hline
\multirow{26}{*}{\parbox{1.0cm}{Deep learning features}}
& Frames+CNN~\cite{Ke2017} (Skel.) & 60.73 & 73.71 & 58.82 & 46.47 & -- \\
& Clips+CNN+Pooling~\cite{Ke2017} (Skel.) & 67.64 & 74.86 & 58.82 & 46.47 & -- \\
& Clips+CNN+Concatenation~\cite{Ke2017} (Skel.) & 71.27 & 78.29 & 61.76 & 53.85 & -- \\
& Clips+CNN+MTLN~\cite{Ke2017} (Skel.) & 73.82 & 79.43 & 67.65 & 54.81 & 2.25 \\
\cline{2-7}
& HPM+AP~\cite{Rahmani2016CVPR} (Depth) & 56.73 & 56.11 & 44.12 & 42.32 & -- \\
& HPM+TM~\cite{Rahmani2016CVPR} (Depth) & 72.00 & {\bf 98.33} & 44.12 & 54.78 & 2.50\\
\cline{2-7}
& 1-layer RNN~\cite{amir2016} (Skel.) & 18.02 & 32.76 & 54.90 & 14.27 & --\\
& 2-layer RNN~\cite{amir2016} (Skel.) & 27.80 & 56.13 & 54.91 & 35.36 & --\\
& 1-layer LSTM~\cite{amir2016} (Skel.) & 62.26 & 67.14 & 61.77 & 50.81 & --\\
& 2-layer LSTM~\cite{amir2016} (Skel.) & 65.33 & 73.72 & 63.24 & 46.78 & --\\
& 1-layer P-LSTM~\cite{amir2016} (Skel.) & 70.50 & 70.86 & 61.76 & {\bf 55.16} & --\\
& 2-layer P-LSTM~\cite{amir2016} (Skel.) & 69.35 & 72.00 & 67.65 & 50.81 & 2.75\\
& \textbf{Our implementation with modified hyperparam. values:} & & & & & \\
& 1-layer LSTM (8 segments, 100 hidden neurons) (Skel.) & 64.75 & 73.14 & 58.82 & 52.58 & --\\
& 2-layer P-LSTM (10 segments, 50 hidden neurons) (Skel.) & 66.09 & 75.43 & 67.65 & 50.00 & -- \\
& 2-layer P-LSTM (10 segments, 100 hidden neurons) (Skel.) & 67.43 & 71.43 & 54.41 & 50.32 & -- \\
& 2-layer P-LSTM (20 segments, 50 hidden neurons) (Skel.) & 73.18 & 71.43 & 52.94 & 49.68 & -- \\
& 2-layer P-LSTM (20 segments, 100 hidden neurons) (Skel.) & 70.50 & 71.43 & 58.82 & 53.55 & -- \\
\cline{2-7}
& IndRNN (4 layers)~\cite{shuai2018} (Skel.) & 71.50 & 90.05 & 51.72 & 44.63 & -- \\
& IndRNN (6 layers)~\cite{shuai2018} (Skel.) & 72.91 & 89.53 & 57.03 & 42.66 & -- \\
& \textbf{Our improved results:} & & & & & \\
& IndRNN (4 layers, with {\jpd}) (Skel.) & 76.34 & 82.66 & {\bf 84.69} & 52.09 & -- \\
& IndRNN (6 layers, with{\jpd}) (Skel.) & {\bf 77.47} & 86.88 & 80.16 & 51.34 & 2.00\\
\cline{2-7}
& ST-GCN$^*$~\cite{stgcn2018aaai} (Skel.) & 27.64 (69.09) & 20.00 (77.14) & 23.53 (70.59)  & 22.12 (45.83) & --\\
& \textbf{Our improved results:} & & & & & \\
& ST-GCN$^*$ (with {\jpd}) (Skel.) & 18.18 (64.00) & 54.16 (96.57) & 26.47 (67.65) & 36.54 (70.51) & 4.75\\
\hline
\end{tabular}}
\label{evaluationresults}
\end{center}
\vspace{-1ex}
{\textsuperscript{*}\footnotesize{For ST-GCN, the numbers inside the parentheses denote the top-5 accuracy.}}
\end{table*}

Table~\ref{evaluationresults} summarizes the results for the single-view action recognition on these
datasets. The algorithms with the highest recognition accuracy for the handcrafted feature and deep learning feature categories are highlighted in bold.

Among the methods that use handcrafted features, SCK+DCK outperformed all other methods on all the datasets. The use of RBF kernels to capture higher-order statistics of the data and complexity of action dynamics demonstrates its effectiveness for action recognition. For the 3D Action Pairs and UWA3D Activity datasets, HON4D and HOPC are, respectively, the second top performers.
The poorer performance of HDG was due to noise in the depth sequences which affected the HDG-{\hod} and HDG-{\hodg} features, even though the human subjects were successfully segmented.
In general, HDG-{\hodg} outperformed HDG-{\hod}, and HDG-{\jmv} outperformed HDG-{\jpd}. We also found that concatenating more individual features in HDG (see Table~\ref{evaluationresults}, row HDG-{\hod}+{\hodg}+{\jpd} to row HDG-all features) helped improve action recognition.
We note that our results for HDG are different from those reported in~\cite{Rahmani2014} because we used \mbox{$1\!\times\!1\!\times\!5=5$} cells~\cite{lei2017} instead of $2\!\times\!2\!\times\!5=20$ cells to store the joint motion features. 

For deep learning methods, the 1-layer P-LSTM method (8 video segments, 50 hidden neurons) outperformed others on the UWA3D Activity dataset. In general, a 1-layer LSTM with more hidden neurons performed better than a 1-layer LSTM with fewer neurons, and P-LSTM performed better than the traditional LSTM and RNN (see the last column in Table~\ref{evaluationresults}). The 2-layer P-LSTM (20 video segments, 50 hidden neurons) achieved the best recognition accuracy on the MSRAction3D dataset and HPM+TM outperformed on the 3D Action Pairs dataset. Comparing the last 5 rows of Table~\ref{evaluationresults} for different variants of 2-layer P-LSTM shows that having more video segments and/or hidden neurons does not guarantee better performance. The reasons are: (i)~more video segments have less averaging effect and so it is likely that noisy video frames with unreliable skeletal information would be used for feature representation; (ii)~having too many hidden neurons would cause overfitting in the training process. 

Using the {\jpd} features instead of the raw 3D joint coordinates boosts the recognition accuracies for both IndRNN and ST-GCN on almost all the datasets. For the CAD-60 and MSRAction3D datasets, the 4-layer IndRNN and 6-layer IndRNN using {\jpd} features are the top two performers. Compared to using the raw 3D joint coordinates, the improvement due to the use of {\jpd} is 4.56\% for IndRNN (6 layers) on the MSRAction3D dataset and 32.97\% for IndRNN (4 layers) on the CAD-60 dataset.


The last column of Table~\ref{evaluationresults} computed using Eq.~\eqref{eq:rank} shows that SCK+DCK obtains average rank 1 score, followed closely by the 6-layer IndRNN with {\jpd} with average rank 2 score. The ST-GCN method uses a more complex architecture having 9 layers of spatio-temporal graph convolutional operators, which require a large dataset for training. As all the datasets in Table~\ref{evaluationresults} are quite small, its poor performance (average rank 4.75 score) is not unexpected.

\subsection{NTU RGB+D Dataset}

Table~\ref{nturesults}~summarizes the evaluation results for cross-subject and cross-view action recognition on the NTU RGB+D dataset, where methods are grouped into the handcrafted feature and deep learning feature categories. Top performing methods are highlighted in bold.

Among the methods using handcrafted features, SCK+DCK performed the best for both cross-subject action recognition and cross-view action recognition. Similarly to results on the four datasets shown in Table~\ref{evaluationresults}, combining more individual features in HDG resulted in higher recognition accuracy for both the cross-subject and cross-view action recognition. 

Among deep learning methods, ST-GCN and the 6-layer IndRNN combined with the {\jpd} features outperformed other deep learning methods in both cross-subject and cross-view action recognition. In particular, with {\jpd}, ST-GCN became the top performer (achieving 83.36\%) for cross-subject action recognition and IndRNN (6 layers) achieved the highest accuracy (89.0\%) for cross-view action recognition. Compared to using the raw 3D skeleton joint coordinates, using the {\jpd} features helps improve the recognition accuracies of both IndRNN and ST-GCN. For example, with {\jpd} features, both the top-1 and top-5 accuracies of ST-GCN increased by 1.79\% and 0.61\% in the cross-subject experiment. 

For the other deep learning methods, the recognition accuracies of 2-layer RNN, 2-layer LSTM and 2-layer P-LSTM are higher than those of 1-layer RNN, 1-layer LSTM and 1-layer P-LSTM. Similar to the results in Table~\ref{evaluationresults}, P-LSTM performed better than the traditional LSTM and RNN. HPM+AP and HPM+TM, on the other hand, did not perform so well. The reason is that both of these methods were trained given only 339 representative human poses from a human pose dictionary whereas the 60 action classes of NTU RGB+D dataset include many more human poses of higher complexity. 


\begin{table}[t!]
\caption{Comparison of average recognition accuracies (percentage) for both cross-subject and cross-view action recognition on the NTU RGB+D Dataset.}
\begin{center}
\resizebox{0.48\textwidth}{!}{\begin{tabular}{ c | l | c | c |}
\cline{1-4}
\multicolumn{1}{c|}{~} & {\bf Method} & {\bf Cross-subject} & {\bf Cross-view} \\
\hline
\multirow{14}{*}{\parbox{1.0cm}{Hand-crafted features}}
& HON4D~\cite{Oreifej2013} (Depth) & 30.6 & 7.3\\
\cline{2-4}
& HOPC~\cite{Rahmani2016} (Depth) & 40.3 & 30.6\\
\cline{2-4}
& LARP-SO-FTP~\cite{Vemulapalli2016} (Skel.) & 52.1 & 53.4\\
\cline{2-4}
& HDG-{\hod}~\cite{lei2017} (Depth) & 20.1 & 13.5\\
& HDG-{\hodg}~\cite{lei2017} (Depth) & 23.0 & 25.2\\
& HDG-{\jpd}~\cite{lei2017} (Skel.) & 27.8 & 35.9\\
& HDG-{\jmv}~\cite{lei2017} (Skel.) & 38.1 & 50.0\\
& HDG-{\hod}+{\hodg}~\cite{lei2017} (Depth) & 24.6 & 26.5\\
& HDG-{\jpd}+{\jmv}~\cite{lei2017} (Skel.) & 39.7 & 51.9\\
& HDG-{\hod}+{\hodg}+{\jpd}~\cite{lei2017} (Depth + Skel.) & 29.4 & 38.8\\
& HDG-{\hod}+{\hodg}+{\jmv}~\cite{lei2017} (Depth + Skel.) & 39.0 & 57.0\\
& HDG-{\hodg}+{\jpd}+{\jmv}~\cite{lei2017} (Depth + Skel.) & 41.2 & 57.2\\
& HDG-all features~\cite{lei2017} (Depth + Skel.) & 43.3 & 58.2\\
\cline{2-4}
& SCK+DCK~\cite{piotr2016} (Skel.) & {\bf 72.8} & {\bf 74.1}\\
\hline
\multirow{22}{*}{\parbox{1.0cm}{Deep learning features}}
& Frames+CNN~\cite{Ke2017} (Skel.) & 75.7 & 79.6\\
& Clips+CNN+Concatenation~\cite{Ke2017} (Skel.) & 77.1 & 81.1\\
& Clips+CNN+Pooling~\cite{Ke2017} (Skel.) & 76.4 & 80.5\\
& Clips+CNN+MTLN~\cite{Ke2017} (Skel.) & 79.6 & 84.8\\
& ${\text{Clips+CNN+MTLN}^\ddagger}$~\cite{Ke2017} (Skel.) & 79.54 & 84.70\\
\cline{2-4}
& HPM+AP~\cite{Rahmani2016} (Depth) & 40.2 & 42.2 \\
& HPM+TM~\cite{Rahmani2016} (Depth) & 50.1 & 53.4 \\
\cline{2-4}
& 1-layer RNN~\cite{amir2016} (Skel.) & 56.0 & 60.2\\
& 2-layer RNN~\cite{amir2016} (Skel.) & 56.3 & 64.1\\
& 1-layer LSTM~\cite{amir2016} (Skel.) & 59.1 & 66.8\\
& 2-layer LSTM~\cite{amir2016} (Skel.) & 60.7 & 67.3\\
& 1-layer P-LSTM~\cite{amir2016} (Skel.) & 62.1 & 69.4\\
& 2-layer P-LSTM~\cite{amir2016} (Skel.) & 62.9 & 70.3\\
& ${\text{2-layer P-LSTM}^\ddagger}$~\cite{amir2016} (Skel.) & 63.02 & 70.39\\
\cline{2-4}
 & IndRNN (4 layers)~\cite{shuai2018} (Skel.) & 78.6 & 83.8 \\
 & IndRNN (6 layers)~\cite{shuai2018} (Skel.) & 81.8 & 88.0 \\
 & \textbf{Our improved results:} & & \\
  & IndRNN (4 layers, with {\jpd})(Skel.) & 79.5 &  84.5 \\
 & IndRNN (6 layers, with {\jpd})(Skel.) & 83.0 & {\bf 89.0} \\
\cline{2-4} 
& ST-GCN$^*$~\cite{stgcn2018aaai} (Skel.) & 81.57 (96.85) & 88.76 (98.83) \\
& \textbf{Our improved results:} & & \\
& ST-GCN$^*$ (with {\jpd}) (Skel.) &  {\bf 83.36} (97.46)  &  88.84 (98.87)  \\
\hline
\end{tabular}}
\label{nturesults}
\end{center}
\vspace{-1ex}
\textsuperscript{$\ddagger$}\footnotesize{Our implementations for reproducing original authors' experiment results.} \\
\textsuperscript{$^*$}\footnotesize{For ST-GCN, the numbers inside the parentheses denote the top-5 accuracy.}
\end{table}

\subsection{UWA3D Multiview Activity II Dataset}
The ten algorithms were compared on the UWA3D Multiview Activity II dataset using cross-view action recognition. Table~\ref{uwa3dmresults}~summarizes the results. The top 2 action recognition algorithms are highlighted in bold in each column for the handcrafted feature and deep learning feature categories.

For the methods using handcrafted features, the HDG-all features performed the best for cross-view action recognition (the last column in Table~\ref{uwa3dmresults}) followed by other HDG variants that use skeletons and depth. 
Among skeleton-only methods, HDG-{\jpd}+{\jmv} was the second best performer followed by SCK+DCK, HDG-{\jmv}, and LARP-SO-FTP, which all performed better than depth-based features such as HON4D, HDG-{\hod}, HDG-{\hodg} and HDG-{\hod}+{\hodg}. According to the table, using one or both skeleton-based features (HDG-{\jpd} and/or HDG-{\jmv}) in HDG  improved  its results.


For deep learning methods, HPM+TM and HPM+AP achieved the highest results. Although Clips+CNN+MTLN performed better than other `Clips' variants, it did not perform as good as HPM+TM and HPM+AP, due to the limited number of video samples in the dataset. 
P-LSTM performed better than the traditional LSTM and RNN. We noticed that stacking more layers for IndRNN or using {\jpd} features for both IndRNN and ST-GCN did not help increase the recognition accuracy. This is due to the lack of representative training videos (compared to the results on the NTU RGB+D dataset in cross-view action recognition).

\begin{table*}[t!]
\begin{center}
\caption{Comparison of average recognition accuracies (percentage) for cross-view action recognition on the UWA3D Multiview Activity II dataset.}
\resizebox{\textwidth}{!}{\begin{tabular}{ c | l | c  c  c  c  c c  c  c  c  c c  c  c |}
\cline{1-15}
\multicolumn{1}{c|}{~} & Training view & \multicolumn{2}{c}{$V_1$ \& $V_2$} & \multicolumn{2}{c}{$V_1$ \& $V_3$} & \multicolumn{2}{c}{$V_1$ \& $V_4$} & \multicolumn{2}{c}{$V_2$ \& $V_3$} & \multicolumn{2}{c}{$V_2$ \& $V_4$} & \multicolumn{2}{c}{$V_3$ \& $V_4$} & Average\\
\cline{2-14}
\multicolumn{1}{c|}{~} & Testing view & $V_3$ & $V_4$ & $V_2$ & $V_4$ & $V_2$ & $V_3$ & $V_1$ & $V_4$ & $V_1$ & $V_3$ & $V_1$ & $V_2$ & {}\\
\hline
\multirow{18}{*}{\parbox{1.0cm}{Hand-crafted features}}
& HON4D~\cite{Oreifej2013} (Depth) & 31.1 & 23.0 & 21.9 & 10.0 & 36.6 & 32.6 & 47.0 & 22.7 & 36.6 & 16.5 & 41.4 & 26.8 & 28.9\\
\cline{2-15}
& HOPC~\cite{Rahmani2016} (Depth) & 25.7 & 20.6 & 16.2 & 12.0 & 21.1 & 29.5 & 38.3 & 13.9 & 29.7 & 7.8 & 41.3 & 18.4 & 22.9\\
& ${\text{Holistic HOPC}^*}$ ~\cite{Rahmani2016} (Depth) & 32.3 & 25.2 & 27.4 & 17.0 & 38.6 & 38.8 & 42.9 & 25.9 & 36.1 & 27.0 & 42.2 & 28.5 & 31.8\\
& ${\text{Local HOPC+STK-D}^*}$ ~\cite{Rahmani2016} (Depth) & 52.7 & 51.8 & {\bf 59.0} & {\bf 57.5} & 42.8 & 44.2 & 58.1 & 38.4 & 63.2 & 43.8 & 66.3 & 48.0 & 52.2\\
\cline{2-15}
& ${\text{LARP-SO-logarithm map}}$ \cite{Vemulapalli2016} (Skel.) & 48.2 & 47.4 & 45.5 & 44.9 & 46.3 & 52.7 & 62.2 & 46.3 & 57.7 & 45.8 & 61.3 & 40.3 & 49.9\\ 
& ${\text{LARP-SO-unwrapping while rolling}}$ \cite{Vemulapalli2016} (Skel.) & 50.4 & 45.7 & 44.0 & 44.5 & 40.8 & 49.6 & 57.4 & 44.4 & 57.6 & 47.4 & 59.2 & 40.8 & 48.5\\
& ${\text{LARP-SO-FTP}}$ \cite{Vemulapalli2016} (Skel.) &54.9 & 55.9 &50.0 &54.9& 48.1 & 56.0 & 66.5 & {\bf 57.2} & 62.5 & 54.0 & 68.9 & 43.6 & 56.0\\ 
\cline{2-15}
& HDG-{\hod}~\cite{lei2017} (Depth) & 22.5 & 17.4 & 12.5 & 10.0 & 19.6 & 20.4 & 26.7 & 13.0 & 18.7 & 10.0 & 27.9 & 17.2 & 18.0\\
& HDG-{\hodg}~\cite{lei2017} (Depth) & 26.9 & 34.2 & 20.3 & 18.6 & 34.7 & 26.7 & 41.0 & 29.2 & 29.4 & 11.8 & 40.7 & 28.8 & 28.5\\
& HDG-{\jpd}~\cite{lei2017} (Skel.) & 36.3 & 32.4 & 31.8 & 35.5 & 34.4 & 38.4 & 44.2 & 30.0 & 44.5 & 33.7 & 44.4 & 34.0 & 36.6\\
& HDG-{\jmv}~\cite{lei2017} (Skel.) & 57.2 & 59.3 & {\bf 59.3} & 54.3 & 56.8 & 50.6 & 63.4 & 52.4 & 65.7 & 53.7 & 67.7& 56.9 & 58.1\\
& {HDG-{\hod}+{\hodg}}~\cite{lei2017} (Depth) & 26.6 & 33.6 & 17.9 & 19.3 & 34.4 & 26.2 & 40.5 & 27.6 & 28.6 & 11.6 & 38.4 & 29.0 & 27.8\\
& {HDG-{\jpd}+{\jmv}}~\cite{lei2017} (Skel.) & {\bf 61.0} & 61.8 & {\bf 59.3} & 56.0 & 60.0 & 57.4 & 68.8 & 54.2 & {\bf 71.1} & {\bf 57.2} & 69.7 & 59.0 & {\bf 61.3}\\
& {HDG-{\hod}+{\hodg}+{\jpd}}~\cite{lei2017} (Depth + Skel.) & 31.0 & 43.5 & 25.7 & 21.4 & 45.9 & 31.1 & 53.2 & 35.7 & 38.0 & 11.6 & 49.7 & 38.3 & 35.4\\
& {HDG-{\hod}+{\hodg}+{\jmv}}~\cite{lei2017} (Depth + Skel.) & 59.0 & {\bf 62.2} & 58.1 & 52.0 & 62.5 & 57.1 & 66.0 & 54.2 & 67.7 & 52.7 & 70.3 & {\bf 61.1} & 60.2\\
& {HDG-{\hodg}+{\jpd}+{\jmv}}~\cite{lei2017} (Depth + Skel.) & 58.2 & 61.8 & 54.8 & 47.6 & {\bf 63.5} & {\bf 58.7} & {\bf 69.0} & 52.3 & 64.9 & 47.1 & 67.2 & 59.4 & 58.7\\
& {HDG-all features}~\cite{lei2017} (Depth + Skel.) & {\bf 60.9} & {\bf 64.3} & 57.9 & 54.6 & {\bf 62.6} & {\bf 59.2} & 68.9 & 55.8 & 69.8 & {\bf 55.2} & {\bf 71.8} & {\bf 62.6} & {\bf 61.9}\\
\cline{2-15}
& SCK+DCK~\cite{piotr2016} (Skel.) & 56.0 & 61.8 & 50.0 & {\bf 61.8} & 54.1 & 56.7 & {\bf 71.4} & {\bf 56.9} & {\bf 72.9} & 49.6 & {\bf 71.7} & 44.0 & 58.9 \\
\hline
\multirow{22}{*}{\parbox{1.0cm}{Deep learning features}}
& {Frames+CNN}~\cite{Ke2017} (Skel.) & 30.4 & 29.8 & 27.8 & 23.8 & 31.7 & 27.2 & 39.2 & 27.8 & 31.4 & 23.2 & 38.8 & 28.6 & 30.0\\
& {Clips+CNN+Pooling}~\cite{Ke2017} (Skel.) & 31.6 & 33.3 & 30.6 & 27.8 & 35.3 & 29.6 & 44.3 & 31.3 & 35.3 & 23.2 & 43.1 & 31.0 & 33.0\\
& {Clips+CNN+Concatenation}~\cite{Ke2017} (Skel.) & 33.2 & 36.9 & 32.5 & 29.8 & 36.9 & 31.2 & 46.3 & 31.0 & 39.2 & 23.6 & 45.5 & 31.7 & 34.8 \\
& {Clips+CNN+MTLN}~\cite{Ke2017} (Skel.) & 36.4 & 38.9 & 34.1 & 30.6 & 37.7 & 33.2 & 46.7 & 31.3 & 38.8 & 25.6 & 49.8 & 33.7 & 36.4 \\
\cline{2-15}
& {HPM+AP}~\cite{Rahmani2016CVPR} (Depth) & {\bf 68.3} & 51.7 & {\bf 60.2} & {\bf 62.2} & 38.7 & {\bf 50.0} & {\bf 58.7} & 37.8 & {\bf 70.6} & {\bf 61.6} & {\bf 74.0} & {\bf 55.6} & {\bf 57.5}\\
& {HPM+TM}~\cite{Rahmani2016CVPR} (Depth) & {\bf 81.7} & {\bf 76.4} & {\bf 74.1} & {\bf 78.7} & {\bf 57.9} & {\bf 69.4} & {\bf 75.8} & {\bf 62.9} & {\bf 81.4} & {\bf 79.9} & {\bf 83.3} & {\bf 73.7} & {\bf 74.6}\\
\cline{2-15}
& {1-layer RNN}~\cite{amir2016} (Skel.) & 11.4 & 11.1 & 10.0 & 14.9 & 11.9 & 13.3 & 11.1 & 15.2 & 10.3 & 12.2 & 9.8 & 12.4 & 12.0\\
& {2-layer RNN}~\cite{amir2016} (Skel.) & 23.4 & 21.6 & 27.3 & 22.7 & 26.3 & 20.3 & 21.7 & 19.5 & 20.7 & 24.8 & 27.0 & 21.5 & 23.1\\
& {1-layer LSTM}~\cite{amir2016} (Skel.) & 28.9 & 12.8 & 19.2 & 15.5 & 20.0 & 33.1 & 48.2 & 16.5 & 43.3 & 15.5 & 38.7 & 16.4 & 25.7\\
& {2-layer LSTM}~\cite{amir2016} (Skel.) & 29.7 & 16.3 & 20.4 & 12.8 & 26.8 & 30.5 & 53.0 & 11.0 & 37.0 & 17.1 & 47.4 & 20.0 & 26.8\\
& {1-layer P-LSTM}~\cite{amir2016} (Skel.) & 26.8 & 23.5 & 22.4 & 22.7 & 24.4 & 31.3 & 49.6 & 20.3 & 38.3 & 19.5 & 45.9 & 17.6 & 28.5\\
& {2-layer P-LSTM}~\cite{amir2016} (Skel.) & 30.1 & 24.7 & 25.2 & 21.5 & 24.4 & 37.4 & 51.0 & 22.7 & 43.1 & 21.1 & 47.8 & 17.2 & 30.5\\
& \textbf{Our implementation with modified hyperparam. values:} & & & & & & & & & & & & &\\
& {2-layer P-LSTM} (10 segments, 50 hidden neurons) (Skel.) & 23.5 & 24.3 & 22.4 & 25.5 & 25.6 & 30.9 & 48.6 & 21.1 & 41.5 & 19.5 & 43.9 & 15.6 & 28.5\\
& {2-layer P-LSTM} (10 segments, 100 hidden neurons) (Skel.) & 21.1 & 22.7 & 24.8 & 21.1 & 22.4 & 33.7 & 47.8 & 21.1 & 45.1 & 22.0 & 49.4 & 17.6 & 29.1\\
& {2-layer P-LSTM} (20 segments, 50 hidden neurons) (Skel.) & 25.2 & 18.3 & 24.8 & 21.5 & 26.4 & 30.5 & 49.0 & 19.1 & 41.5 & 24.4 & 43.5 & 14.8 & 28.3\\
& {2-layer P-LSTM} (20 segments, 100 hidden neurons) (Skel.)  & 27.6 & 24.3 & 24.8 & 21.9 & 28.4 & 32.9 & 44.3 & 24.3 & 44.3 & 20.3 & 45.9 & 15.6 & 29.6\\
\cline{2-15}
& {IndRNN (4 layers)}~\cite{shuai2018} (Skel.) & 34.3 & {\bf 53.8} & 35.2 & 42.5 & {\bf 39.1} & 38.9 & 49.2 & {\bf 42.5} & 46.0 & 27.1 & 48.6 & 30.9 & 40.7\\
& {IndRNN (6 layers)}~\cite{shuai2018} (Skel.) & 30.7 & 47.2 & 32.2 & 36.0 & 38.8 & 35.4 & 44.5 & 37.9 & 40.6 & 23.9 & 39.2 & 25.2 & 36.0\\
& {IndRNN (4 layers, with {\jpd})} (Skel.) & 33.5 & 40.2 & 26.9 & 40.9 & 30.4 & 41.1 & 50.7 & 36.7 & 46.1 & 24.6 & 49.0 & 23.0 & 36.9\\
& {IndRNN (6 layers, with {\jpd})} (Skel.) & 29.4 & 36.3 & 23.9 & 38.2 & 26.0 & 36.8 & 45.4 & 33.0 & 41.2 & 20.5 & 44.7 & 18.6 & 32.8\\
\cline{2-15}
& ST-GCN~\cite{stgcn2018aaai} (Skel.) & 36.4 & 26.2 & 20.6 & 30.2 & 33.7 & 22.4 & 43.1 & 26.6 & 16.9 & 12.8 & 26.3 & 36.5 & 27.6 \\
& ST-GCN (with {\jpd}) (Skel.) & 29.6 & 21.8 & 15.5 & 19.1 & 17.1 & 18.8 & 35.3 & 14.3 & 31.0 & 14.8 & 13.3 & 15.5 & 20.5 \\
\hline
\end{tabular}}
\label{uwa3dmresults}
\end{center}
\vspace{-1ex}
{\textsuperscript{*}\footnotesize{This result is obtained from \cite{Rahmani2016} for comparison.}}
\end{table*}

\begin{figure}[t!]
\centering
\includegraphics[width=\linewidth,viewport=20 0 650 615, clip=true]{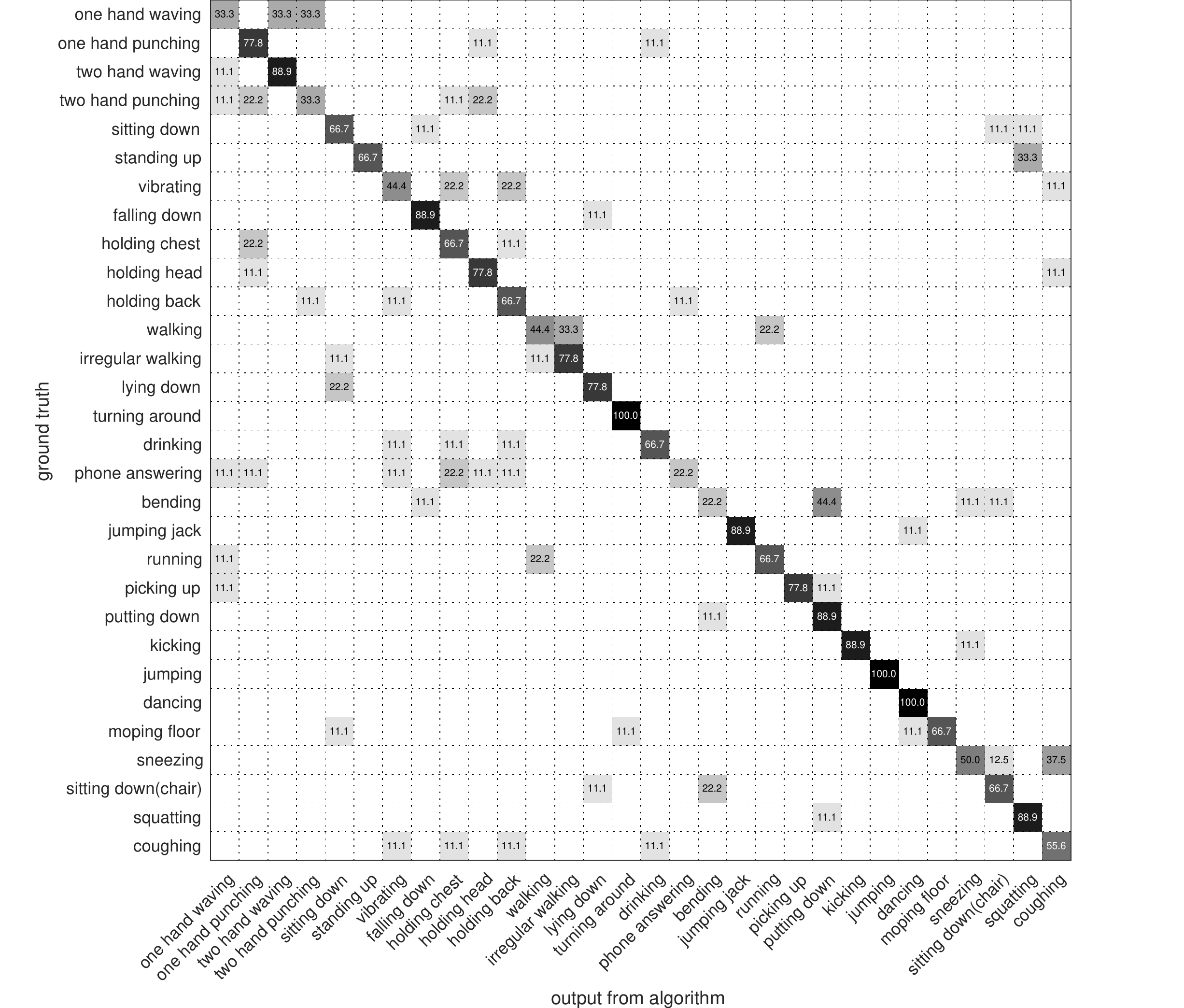}
\caption{Confusion matrix for HDG-all features on the UWA3D Multiview Activity II dataset when $V_3$ and $V_4$ were used for training and $V_1$ was used for testing. For each action class along the diagonal, the darker is the colour, the higher is the recognition accuracy.}
\label{uwamultiviewv3v4v1}
\end{figure}

Fig.~\ref{uwamultiviewv3v4v1} shows a confusion matrix for the HDG-all representation on the UWA3D Multiview Activity II dataset when $V_3$ and $V_4$ views were used for training and $V_1$ was used for testing. According the figure, the algorithm was confused by a few actions which have similar motion trajectories. For example, {\it one hand waving} is similar to {\it two hand waving} and {\it two hand punching}; {\it walking} is very similar to {\it irregular walking}; {\it bending} and {\it putting down} have very similar appearance; {\it sneezing} and {\it coughing} are very similar actions.

\section{Discussions}
\label{sec:discussion}

\subsection{Single-view versus cross-view} Table~\ref{overallresults} summarizes the performance of each algorithm for both single-view action recognition and cross-view action recognition. One can observe from the table that some algorithms such as HON4D, LARP-SO-FTP, Clips+CNN+MTLN, P-LSTM, and SCK+DCK performed better for single-view action recognition but performed slightly worse in cross-view action recognition. For cross-subject action recognition, SCK+DCK outperformed all other algorithms; for cross-view action recognition, HDG-all, SCK+DCK and our improved IndRNN with {\jpd} features all performed well (having the same average rank values). 

\begin{figure}[t!]
\centering
\includegraphics[width=\linewidth]{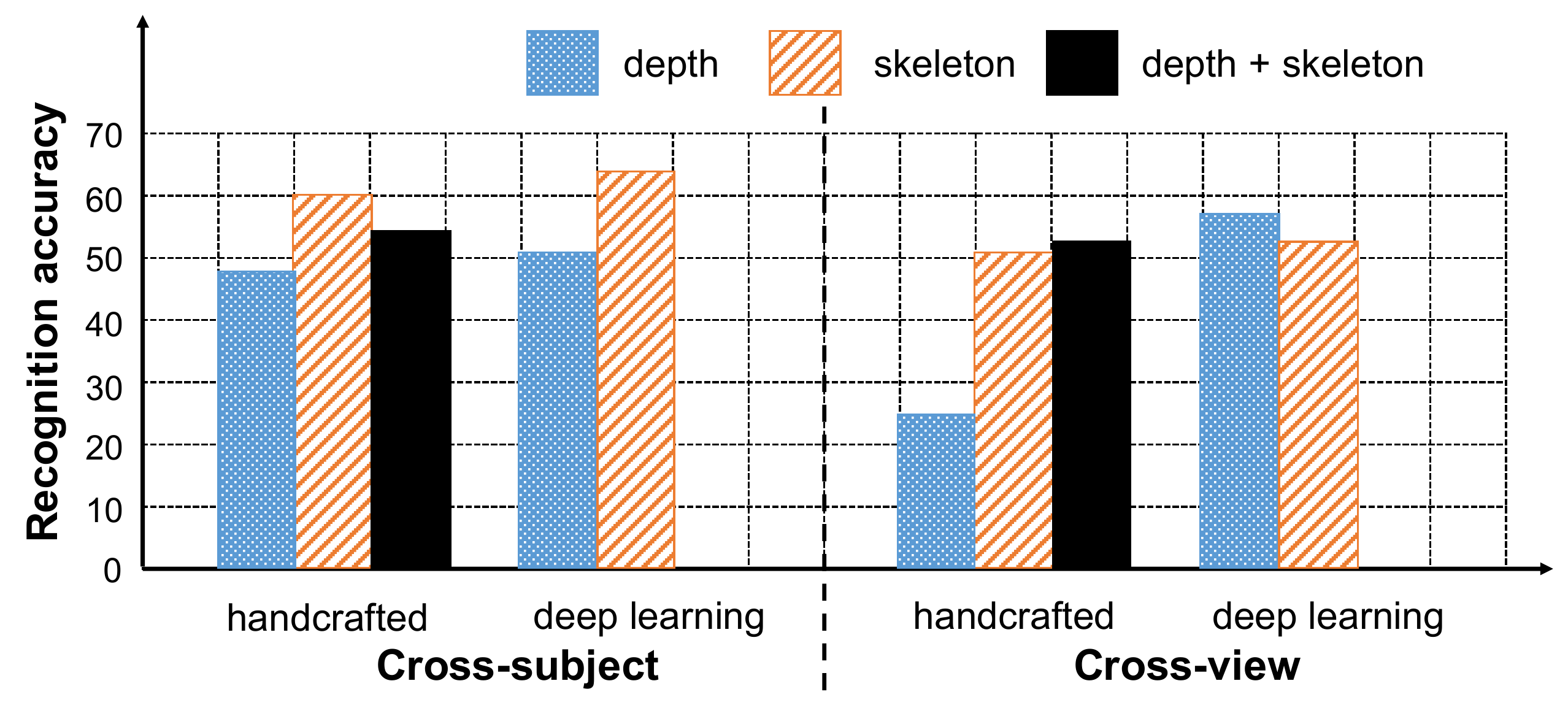} \\
\footnotesize{~~~~~(a) \hspace{3cm} ~~~~~(b)}
\caption{The average recognition accuracy (in percentage)
of methods using handcrafted and deep learning features for cross-subject and cross-view recognition.
Numbers of methods using handcrafted (i)~depth-based features: 7; 
(ii)~skeleton-based features: 7; 
(iii)~depth+skeleton-based features: 4. 
Numbers of methods using deep learning (i)~depth-based features: 2; 
(ii)~skeleton-based features: 20; (iii)~depth+skeleton-based features: 0
(see Tables~\ref{evaluationresults}--\ref{uwa3dmresults}).
} 
\label{new_comparison}
\end{figure}

Fig.~\ref{new_comparison} 
summarizes the average accuracy of all the algorithms grouped under the handcrafted and deep learning feature categories. 
On average, we found that the recognition accuracy for cross-view action recognition is lower than cross-subject action recognition using handcrafted features, with the poorest performance from depth-based features. Combining both depth- and skeleton-based features together helps improve cross-view action recognition and gives a similar performance for cross-subject action recognition. Overall, the performance on cross-view recognition is lower than cross-subject recognition, 
except for the deep learning depth-based feature methods. It should be noted that there are only two methods (HPM+AP and HPM+TM) falling into this category and they performed well on the UWA3D Multiview Activity II dataset and reasonably well on the NTU RGB+D dataset. 

\begin{table}[t!]
\caption{The average recognition accuracies / [AVRank] of all the ten algorithms for both cross-subject and cross-view action recognition.}
\begin{center}
\resizebox{0.48\textwidth}{!}{\begin{tabular}{| l | c | c |}
\hline
 Algorithms & ${\text {Cross-subject}^{\dagger}}$ & ${\text {Cross-view}^{\ddagger}}$ \\ 
\hline
\hline
HON4D~\cite{Oreifej2013} (Depth) & 52.8 / [3.6] & 18.1 / [5.0]\\
\hline
HDG-all~\cite{lei2017} (Depth + Skel.) & 56.6 / [3.8] & 60.1 / {\bf [1.5]}\\
\hline
HOPC~\cite{Rahmani2016} (Depth) & 55.9 / [3.4] & 41.4 / [4.0]\\
\hline
LARP-SO-FTP~\cite{Vemulapalli2016} (Skel.) & 65.0 / [2.4] & 54.7 / [3.0]\\
\hline
SCK+DCK~\cite{piotr2016} (Skel.) & 78.7 / {\bf [1.0]} & 66.5 / {\bf [1.5]} \\
\hline
\hline
HPM+TM~\cite{Rahmani2016CVPR} (Depth) & 58.7 / [3.0] & 64.0 / [3.0]\\
\hline
Clips+CNN+MTLN~\cite{Ke2017} (Skel.) & 74.3 / [2.4] & 60.6 / [3.0]\\
\hline
P-LSTM~\cite{amir2016} (Skel.) & 64.0 / [3.0] & 50.4 / [4.0]\\
\hline
IndRNN~\cite{shuai2018}(Skel.) & 72.6 / [2.4] & 62.1 / [2.0]\\
IndRNN~with {\jpd} (Skel.) & 77.6 / {\bf [2.0]} & 60.8 / {\bf [1.5]} \\
\hline
ST-GCN~\cite{stgcn2018aaai}(Skel.) & 52.4 / [4.2] & 58.2 / [3.5]\\
ST-GCN~with {\jpd} (Skel.) & 57.7 / [4.0] & 54.7 / [3.0]\\
\hline
\end{tabular}}
\label{overallresults}
\end{center}
\vspace{-1ex}
{\textsuperscript{$\dagger$}\footnotesize{For cross-subject action recognition, the performance of each algorithm is computed by averaging all the recognition accuracies / rank values on the MSRAction3D, 3D Action Pairs, CAD-60, UWA3D Activity and NTU RGB+D datasets.}
\quad
{\textsuperscript{$\ddagger$}\footnotesize{For cross-view action recognition, they are computed over the UWA3D Multiview Activity II and NTU RGB+D datasets.}} 
}\\
\end{table}

\begin{figure*}[t!]
    \centering 
\begin{subfigure}{0.32\textwidth}
  \includegraphics[width=\linewidth]{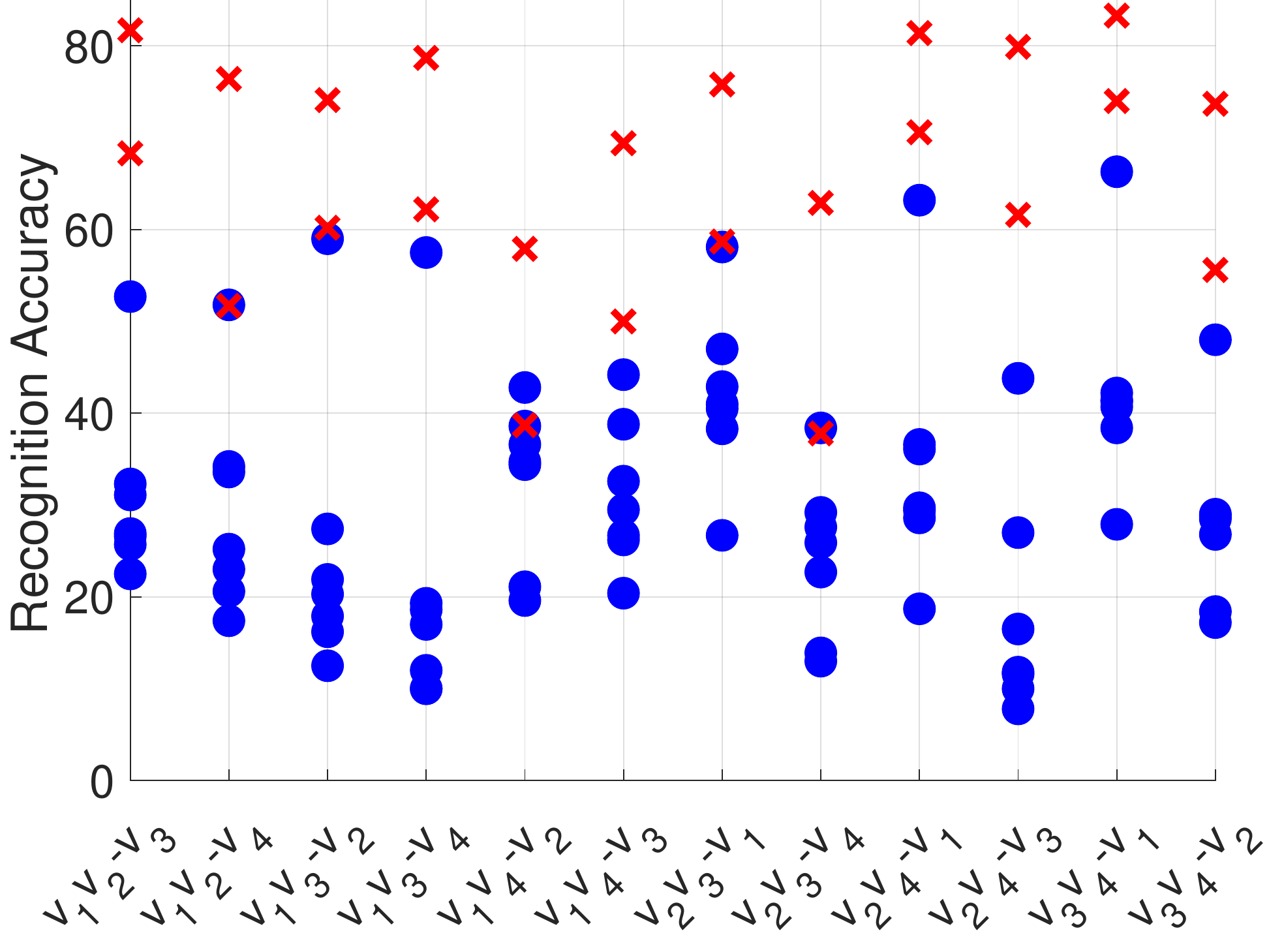}
  \caption{Depth-based features}
  \label{fig_d}
\end{subfigure}\hfil 
\begin{subfigure}{0.32\textwidth}
  \includegraphics[width=\linewidth]{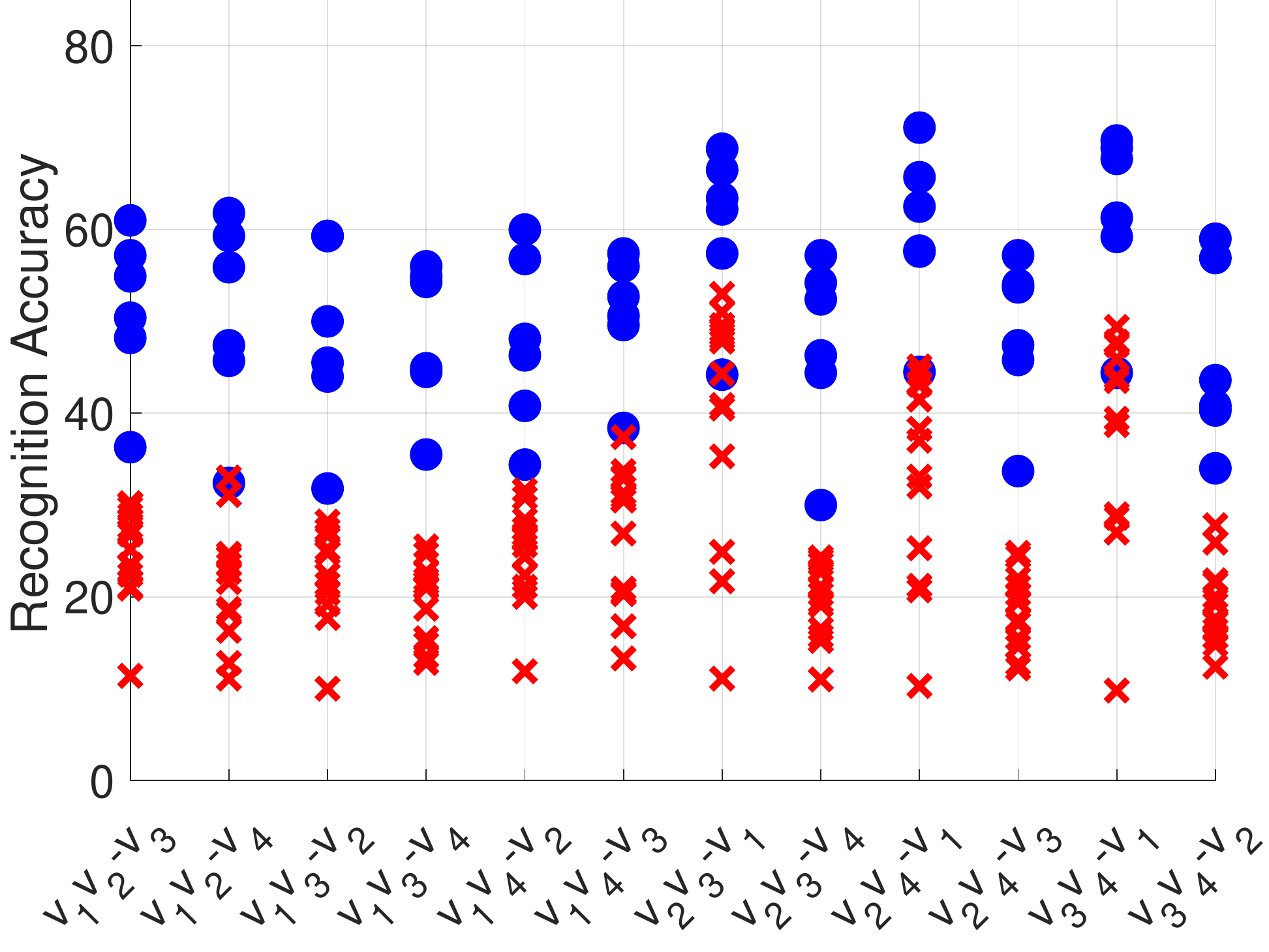}
  \caption{Skeleton-based features}
  \label{fig_s}
\end{subfigure}\hfil 
\begin{subfigure}{0.32\textwidth}
  \includegraphics[width=\linewidth]{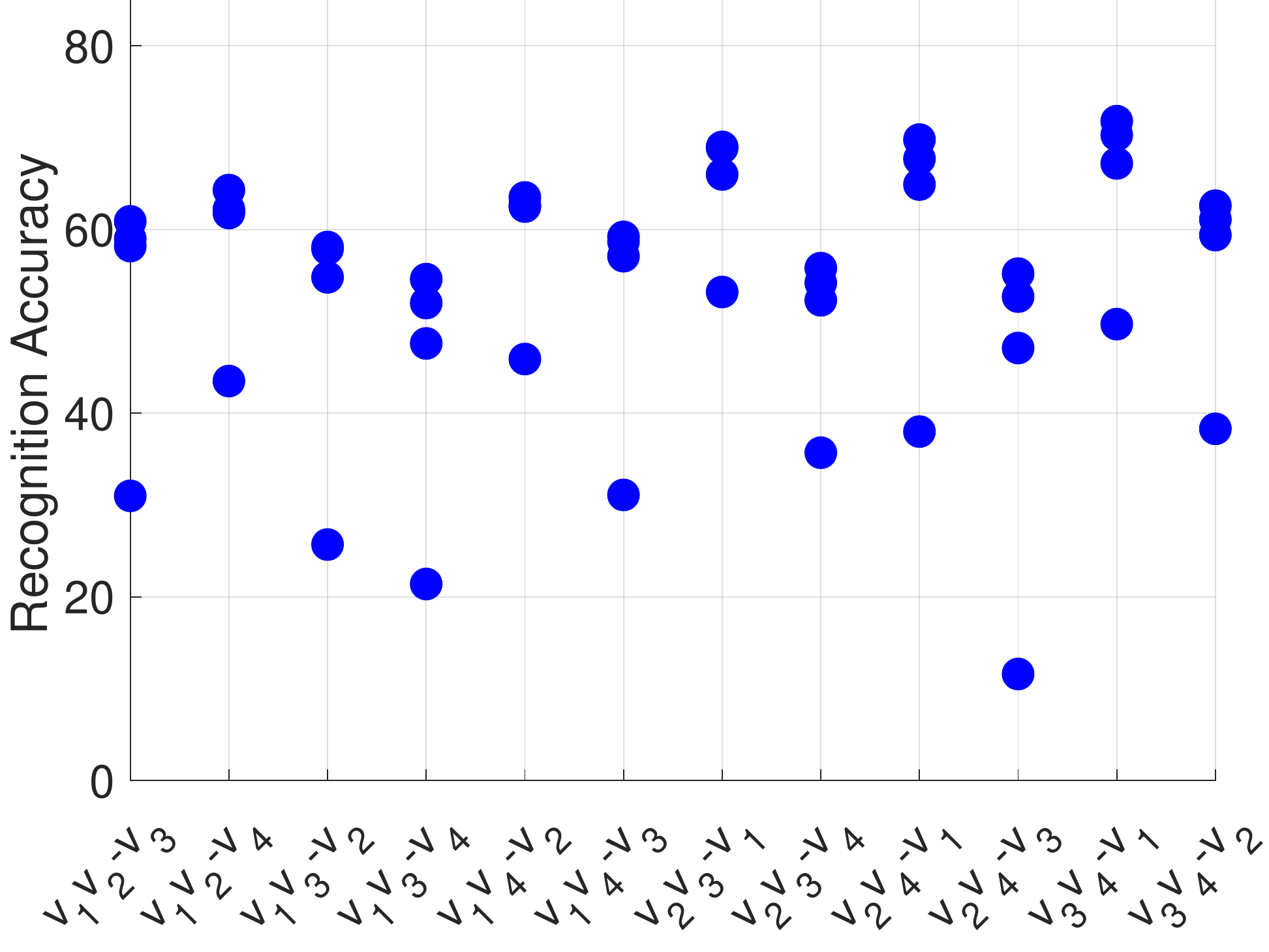}
  \caption{Depth+Skeleton-based features}
  \label{fig_ds}
\end{subfigure}
    \caption{Scatter plots showing the performance of cross-view action recognition on the UWA3D Multiview Activity II dataset. The blue dots and red crosses, respectively, represent methods using handcrafted features and deep learning features. On the horizontal axis of each plot, we use the notation $V_iV_j$-$V_k$ to denote views $i$ and $j$ being used for training and view $k$ being used for testing.}
    \label{scatterplot}
\end{figure*}

\renewcommand{\V}[3]{\mbox{$V_#1V_#2$-$V_#3$}}

\subsection{Influence of camera views in cross-view evaluation} To analyze the influence of using different combinations of camera views in training and testing, we classified all the algorithms in Table~\ref{uwa3dmresults} into 3 feature types: depth-based, skeleton-based, and depth+skeleton-based representations. Scatter plots showing the performance of all the algorithms on these three feature types are given in Fig.~\ref{scatterplot}, where the blue dots and red crosses denote, respectively, algorithms using handcrafted features and deep learning features. 

According to plots in Fig.~\ref{scatterplot}
 the recognition accuracy was high when the left view $V_2$ and right view $V_3$ were used for training and front view $V_1$ for testing (see axis tick mark \V{2}{3}{1}). This is obvious as the viewing angle of the front view $V_1$ is between $V_2$ and $V_3$, as shown in Fig.~\ref{uwa3d_sample}. However, it is surprising that 
the recognition accuracy was even slightly higher for \V{2}{4}{1} and \V{3}{4}{1} (see Figs.~\ref{scatterplot}(b) and~\ref{scatterplot}(c)).

One can also notice that for depth-based methods which use handcrafted features, the recognition accuracy dropped the most when the left view $V_2$ and top view $V_4$ were used for training, and the right view $V_3$ was used for testing. The main reason for this behavior is that the visual appearances of an action in $V_2$ and $V_3$ are different, and it is very difficult to find the same view-invariant features from these views. Other combinations of views, such as \V{1}{3}{4}, \V{1}{3}{2}, and \V{1}{2}{4}, also led to a lower performance (see the distribution of the blue dots in Fig.~\ref{scatterplot}(a)).




\subsection{Depth-based features versus skeleton-based features}
\label{sec:depth_vs_skeleton}
\vspace{0.05cm}
\noindent{\bf Cross-subject action recognition.}
It seems that for the cross-subject action recognition, skeleton-based features outperformed depth-based features for both the handcrafted and deep learning feature categories (Fig.~\ref{new_comparison}(a)). 
However, it is surprising that adding the depth-based features to skeleton-based features actually led to a slight decrease in the action recognition accuracy.
The main reason is 
the background clutter and noise (for instance, see Fig.~\ref{cad60examples}) in the depth sequences, making the depth-based features less representative for robust action recognition. 

On the other hand, 
some skeleton-based algorithms such as LARP-SO-FTP and its variants performed well in human action recognition because they only extracted reliable features from those human body joints which have high confidence values. 
The only exception is the HOPC algorithm which uses depth-based features and which performed better than skeleton-based features such as HDG-{\jpd}+{\jmv} in the cross-subject action recognition (see the average performance in the last column of Table~\ref{evaluationresults}).  One must note that the HOPC algorithm is different from other depth-based algorithms as it treats the depth images as a 3D pointcloud. Such an approach allows the HOPC algorithm to estimate the orientations of the local surface patches on the body of the human subject to more robustly deal with viewpoint changes.



\vspace{0.05cm}
\noindent{\bf Cross-view action recognition.} In Fig.~\ref{scatterplot}(b), for the cross-view action recognition, methods using handcrafted features (blue dots) performed better  than those using deep learning features (red crosses) for skeleton-based features. For depth-based features (Fig.~\ref{scatterplot}(a)), no clear conclusion can be drawn for the handcrafted versus deep learning features. 

Comparing the handcrafted features (blue dots) in both Figs.~\ref{scatterplot}(a) and~\ref{scatterplot}(b), we can see that algorithms using skeleton-based features produced better results than depth-based features. This is evident from Fig.~\ref{scatterplot}(b) as the blue dots are well above the red crosses, whereas in Fig.~\ref{scatterplot}(a) the blue dots are more spread out and they occupy mainly the lower region of the plot.
Our pruning process of the HDG algorithm also proved that skeleton-based features are more robust than the depth-based algorithms in the cross-view action recognition.

For the plot in Fig.~\ref{scatterplot}(c), the algorithms in our experiments did not use deep learning to compute depth+skeleton features, thus no comparison on the performance between using handcrafted and deep learning features 
is possible. 



\subsection{Handcrafted features versus deep learning features} 


As expected, our experiments confirm that
deep learning methods (\eg, Clips+CNN+MTLN, IndRNN, and ST-GCN) performed better on large datasets such as the NTU RGB+D dataset, which has more than 56,000 video sequences, but achieved a lower recognition accuracy on the other smaller datasets. The main reason for this behavior is the reliance of deep learning methods on large amount of data for training to avoid overfitting (in contrast to handcrafted methods). However, most existing action datasets have small numbers of performing subjects and action classes, and very few camera views due to high cost of collecting/annotating data.

%


We also found that the performance of most handcrafted methods is highly dependent on the datasets. For example, HON4D and HOPC  performed better only on some specific datasets, such as MSRAction3D and 3D Action Pairs, but achieved lower performance on the NTU RGB+D dataset. This means that features that have been handcrafted to work on one dataset may not be transferable to another dataset. Moreover, as handcrafted methods prevent overfitting well, they may lack the capacity to learn from big data. With larger benchmark datasets becoming available to the research community, future research trend is more likely to shift to using deep learning features with efforts being put into tuning their hyperparameters.


\subsection{`Quo Vadis, action recognition?'}

Based on the methods evaluated in this paper, we see the following trends in human action recognition. Handcrafted features progressed from global descriptors (\eg, HON4D in 2013) to local descriptors (\eg, HOPC in 2016) and combinations of global and local feature representation (\eg, HDG in 2014). More recent works focus on designing robust 3D human joint-based representations (\eg, LARP-SO-FTP and SCK+DCK in 2016). Moreover, researchers have adopted features extracted from skeleton sequences as they are easier to process and analyze than  depth videos. We also note that deep learning representations are evolving from basic neural networks (\eg, traditional RNN and LSTM) to adapted and/or dedicated networks that rely on a pre-trained network (\eg, HPM+AP, HPM+TM, and Clips+CNN+MTLN). Researchers also modify basic neural network models to improve their learning capability for action recognition (\eg, P-LSTM and IndRNN). Recently, new models such as ST-GCN were devised to robustly model the human actions in large datasets.  Given the surge of large datasets and powerful GPU machines, the newest methods learn representations in an end-to-end manner.

\section{Conclusion}
\label{sec:conclusion}

We have presented, analyzed and compared ten state-of-the-art algorithms in 3D action recognition
on six benchmark datasets. These algorithms cover the use of handcrafted and deep learning features computed from depth and skeleton video sequences. We believe that our comparison results will be useful for future researchers interested in designing robust representations for recognizing human actions from videos captured by the Kinect camera and/or similar action sequence capturing sensors. 

In our study, we found that skeleton-based features are more robust than depth-based features for both cross-subject action recognition and cross-view action recognition. Handcrafted features performed better than deep learning features given smaller datasets. However, deep learning methods achieved very good results if trained on large datasets. 
While accuracy as high as 90\% has been achieved by some algorithms on cross-subject action recognition, the average accuracy on cross-view action recognition is much lower. 

From our literature review and comprehensive experiments, we see that many research papers on human action recognition have already attempted to tackle the challenging issues mentioned in the introduction of the paper. Examples include using a skeleton model or treating depth maps as 3D pointcloud to overcome the changes of viewpoint; using Fourier temporal pyramid or dynamic time warping to deal with different speeds of execution of actions; using depth videos to overcome changes in lighting conditions and eliminate visual appearances  that may be irrelevant to the actions performed.  While these papers all report promising action recognition results, new and more robust action recognition algorithms are still required, given that there is a pressing demand for using these algorithms in real and new environments. 



\section*{Acknowledgments}

We would like to thank the authors of HON4D~\cite{Oreifej2013}, HOPC~\cite{Rahmani2016}, LARP-SO~\cite{Vemulapalli2016}, HPM+TM~\cite{Rahmani2016CVPR}, IndRNN~\cite{shuai2018} and ST-GCN~\cite{stgcn2018aaai} for making their codes publicly available. We thank the ROSE Lab of Nanyang Technological University (NTU), Singapore, for making the NTU RGB+D dataset freely accessible. We would like to thank Data61/CSIRO for providing the computing cluster for running our experiments.

\scriptsize
{
\bibliographystyle{IEEEtranN}
\bibliography{researchbib}}

\vskip -2\baselineskip plus -1fil

\begin{IEEEbiography}
[{\includegraphics[width=1in,height=1.25in,clip,keepaspectratio]{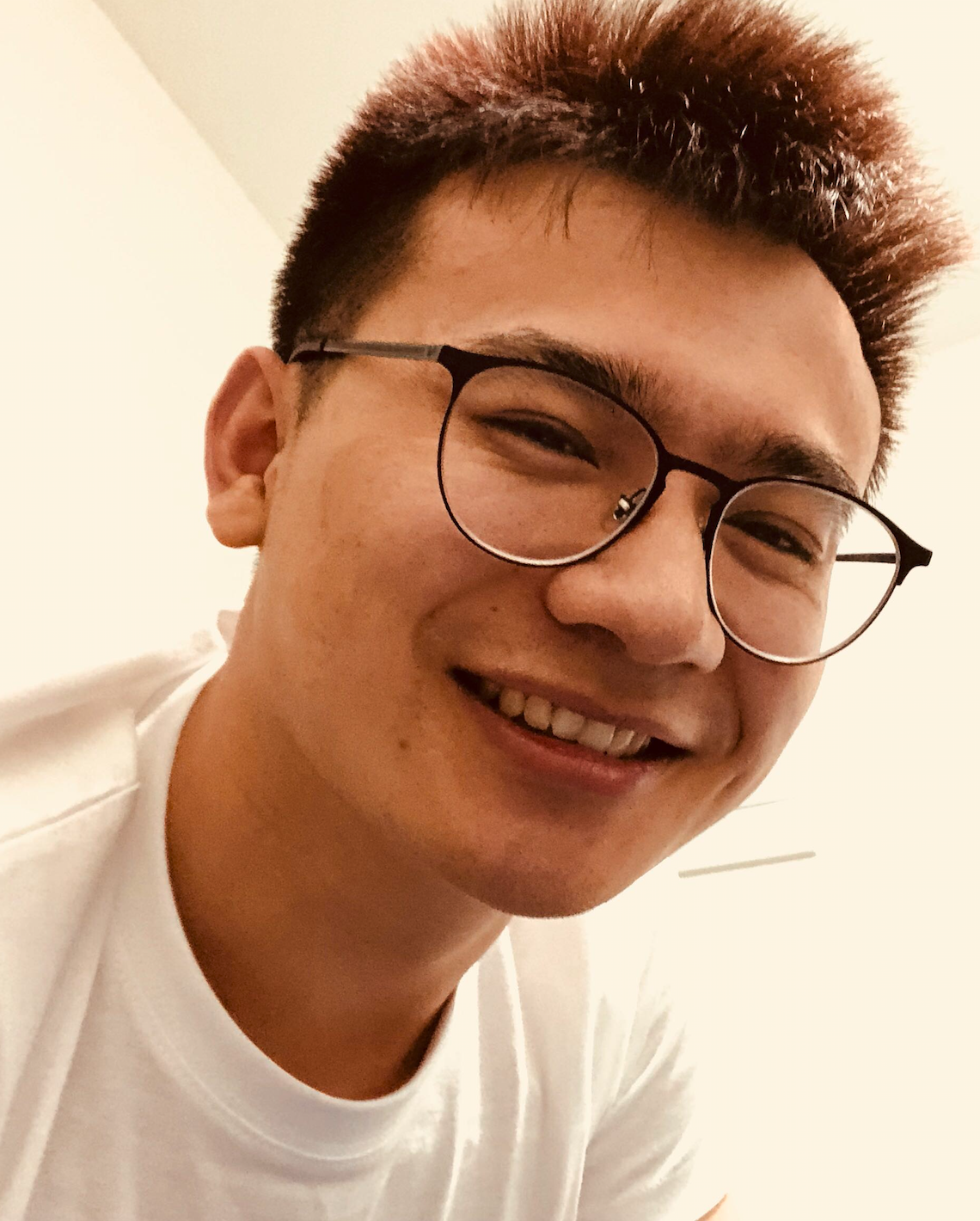}}]{Lei Wang}
received the 
M.E. degree in software engineering from The University of Western Australia (UWA), 
Australia, in 2018.
Since then, he has worked as a Computer Vision Researcher at iCetana Pty Ltd. He was a Visiting Researcher in Machine Learning Research Group at Data61/CSIRO, and he is a Visiting Researcher with the Department of Computer Science and Software Engineering, UWA.
He is also pursuing a Ph.D. degree 
under the supervision of Dr. Piotr Koniusz and A/Prof Du Q. Huynh at the Australian National University (ANU) and Data61/CSIRO. 
His research interests include human action recognition in videos, machine learning and computer vision.
\end{IEEEbiography}

\vskip -2\baselineskip plus -1fil

\begin{IEEEbiography}
[{\includegraphics[width=1in,height=1.25in,clip,keepaspectratio]{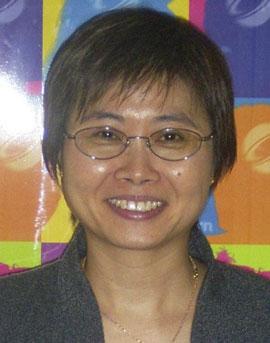}}]{Du Q. Huynh}
received the PhD degree in computer vision from The University of Western Australia, Crawley, Australia, in 1994. Since then, she has worked for the Australian Cooperative Research Centre for Sensor Signal and Information Processing (CSSIP) and Murdoch University. She is currently an associate professor with the Department of Computer Science and Software Engineering, The University of Western Australia. 
Her research interests include visual target tracking, video image processing, machine learning, and pattern recognition. She has previously worked on shape from motion, multiple view geometry, and 3D reconstruction. 
\end{IEEEbiography}

\vskip -2\baselineskip plus -1fil

\begin{IEEEbiography}
[{\includegraphics[trim=50 0 0 0,width=1in,height=1.25in,clip,keepaspectratio]{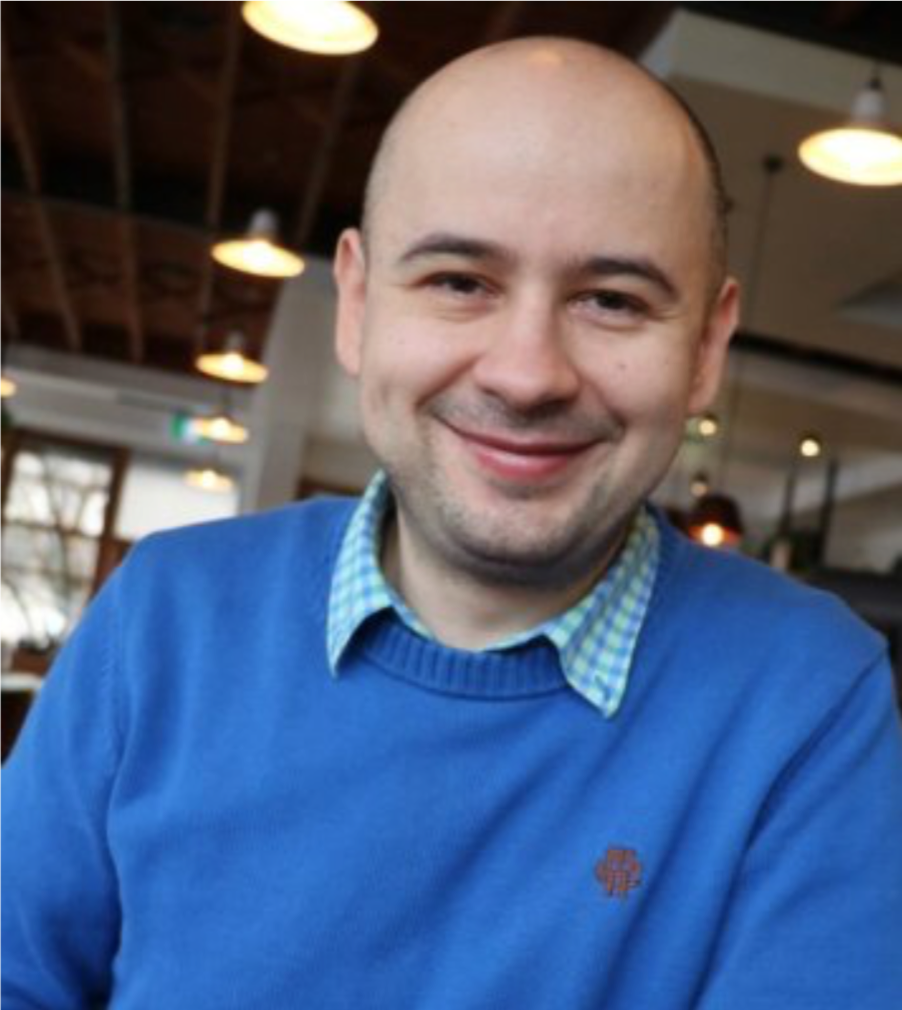}}]{Piotr Koniusz}
A Senior Researcher in Machine Learning Research Group at Data61/CSIRO, formerly known as NICTA, and a Senior Honorary Lecturer at Australian National University (ANU). Previously, he worked as a postdoctoral researcher in the team LEAR, INRIA, France. He received his BSc degree in Telecommunications and Software Engineering in 2004 from the Warsaw University of Technology, Poland, and completed his PhD degree in Computer Vision in 2013 at CVSSP, University of Surrey, UK. His interests include visual categorization, spectral learning on graphs and tensors, kernel methods and deep learning.
\end{IEEEbiography}

\end{document}